\documentclass[final]{cvpr}

\usepackage{times}
\usepackage{epsfig}
\usepackage{graphicx}
\usepackage{amsmath}
\usepackage{amssymb}
\usepackage{xcolor}
\usepackage{amsmath,amsfonts,bm}

\def\eqref#1{equation~\ref{#1}}
\def\1{\bm{1}}

\def\mK{{\bm{K}}}

\def\mP{{\bm{P}}}

\def\mS{{\bm{S}}}

\def\mV{{\bm{V}}}

\DeclareMathAlphabet{\mathsfit}{\encodingdefault}{\sfdefault}{m}{sl}
\SetMathAlphabet{\mathsfit}{bold}{\encodingdefault}{\sfdefault}{bx}{n}
\newcommand{\tens}[1]{\bm{\mathsfit{#1}}}

\def\tF{{\tens{F}}}

\def\tI{{\tens{I}}}

\def\tM{{\tens{M}}}

\def\tO{{\tens{O}}}
\def\tP{{\tens{P}}}

\def\tU{{\tens{U}}}

\def\tW{{\tens{W}}}

\newcommand{\etens}[1]{\mathsfit{#1}}

\def\etD{{\etens{D}}}

\def\etF{{\etens{F}}}

\def\etI{{\etens{I}}}

\def\etM{{\etens{M}}}

\newcommand{\reg}{\lambda}

\usepackage{url}
\usepackage{graphicx} 
\usepackage{multirow}
\usepackage{caption}
\usepackage{wrapfig}
\usepackage{comment}
\usepackage{booktabs} % professional-quality tables
\usepackage{enumitem} % allow set itemize indent 
\usepackage{soul}
\usepackage[normalem]{ulem}

\usepackage[pagebackref=true,breaklinks=true,colorlinks,bookmarks=false]{hyperref}

\def\cvprPaperID{10375} % *** Enter the CVPR Paper ID here
\def\confYear{CVPR 2021}
\title{CoCoNets: Continuous Contrastive 3D Scene Representations}

\makeatletter
\newcommand{\printfnsymbol}[1]{%
  \textsuperscript{\@fnsymbol{#1}}%
}

\author{Shamit Lal\thanks{Equal contribution} , Mihir Prabhudesai\printfnsymbol{1}, Ishita Mediratta, Adam W. Harley, Katerina Fragkiadaki\\
Carnegie Mellon University\\
{\tt\small \{shamitl, mprabhud, imedirat, aharley, katef\} @cs.cmu.edu}
}

\newcommand\blfootnote[1]{%
  \begingroup
  \renewcommand\thefootnote{}\footnote{#1}%
  \addtocounter{footnote}{-1}%
  \endgroup
}

\begin{document}
\maketitle
% \thispagestyle{empty}
% \pagenumbering{gobble}
\blfootnote{
Project page: \href{https://mihirp1998.github.io/project\_pages/coconets/}{https://mihirp1998.github.io/project\_pages/coconets/}}

\newcommand{\Iz}{{\displaystyle \etI}^{(1)}} %
\newcommand{\It}{{\displaystyle \etI}^{(t)}} % 
\newcommand{\In}{{\displaystyle \etI}^{(n)}} % 
\newcommand{\Ino}{{\displaystyle \etI}^{(n+1)}} % 
\newcommand{\IV}{{\displaystyle \etI}^{(V)}} % 
\newcommand{\Ito}{{\displaystyle \etI}^{(t+1)}} % 
\newcommand{\three}{^{\textrm{3D}}}
\newcommand{\two}{^{\textrm{2D}}} 
\newcommand{\ctwo}{\mathcal{T}^{\textrm{2D}}} 
\newcommand{\cthree}{\mathcal{T}^{\textrm{3D}}} 
\newcommand{\btwo}{\mathcal{B}^{\textrm{2D}}} 
\newcommand{\bthree}{\mathcal{B}^{\textrm{3D}}} 

\newcommand{\flowt}{{\displaystyle \tW}^{(t)}}

\newcommand{\Etop}{{C}} % 
\newcommand{\Ebot}{{B}} % 
\newcommand{\losst}{\ell^{(t)}}
\newcommand{\loss}{\mathcal{L}}

\newcommand{\ctx}{\mathcal{T}} % 
\renewcommand{\bot}{\mathcal{B}} % 

% unprojected rgb
\newcommand{\Ut}{{\displaystyle \tU}^{(t)}}
\newcommand{\Uz}{{\displaystyle \tU}^{(1)}}
\newcommand{\Uto}{{\displaystyle \tU}^{(t+1)}} 
\newcommand{\Iht}{\hat{{\displaystyle \tI}}^{(t)}}

\newcommand{\obo}{\textrm{reg}}
\renewcommand{\reg}{\textrm{reg}}
\newcommand{\rUt}{{{\displaystyle \tU}}^{(t)}_{\textrm{reg}}}
\newcommand{\rUz}{{{\displaystyle \tU}}^{(1)}_{\textrm{reg}}}
\newcommand{\rOt}{{{\displaystyle \tO}}^{(t)}_{\textrm{reg}}}
\newcommand{\rOz}{{{\displaystyle \tO}}^{(1)}_{\textrm{reg}}}

% camera pose
\newcommand{\Pt}{{\displaystyle \mP}^{(t)}} 
\newcommand{\Vt}{{\displaystyle \mV}^{(t)}} 
\newcommand{\K}{{\displaystyle \mK}} 
\renewcommand{\S}{{\displaystyle \mS}} 

\newcommand{\Vk}{{\displaystyle \mV}^{(k)}}
\newcommand{\Vz}{{\displaystyle \mV}^{(1)}} 
\newcommand{\Vn}{{\displaystyle \mV}^{(n)}} 
\newcommand{\Vno}{{\displaystyle \mV}^{(n+1)}} 

%\Ft_{\theta_i}

% occupancy
\newcommand{\Ot}{{\displaystyle \tO}^{(t)}}
\newcommand{\Oz}{{\displaystyle \tO}^{(1)}}
\newcommand{\Ct}{{\displaystyle \tP}^{(t)}}
\newcommand{\Cht}{\hat{{\displaystyle \tP}}^{(t)}}

% feature tensor
%\newcommand{\F}{{\displaystyle \tF}} % 
\newcommand{\F}{\mathcal{F}}
\newcommand{\Ft}{{\displaystyle \tF}^{(t)}}
\newcommand{\Fi}{{\displaystyle \tF}^{(i)}}
\newcommand{\Fj}{{\displaystyle \tF}^{(j)}}
\newcommand{\Fno}{{\displaystyle \tF}^{(n+1)}}
\newcommand{\rFt}{{\displaystyle \tF}^{(t)}_{\textrm{reg}}}
\newcommand{\rFto}{{\displaystyle \tF}^{(t+1)}_{\textrm{reg}}}
\newcommand{\rFz}{{\displaystyle \tF}^{(1)}_{\textrm{reg}}}
\newcommand{\Ftti}{{\displaystyle \tF}^{(t)}_{\theta_i}}
\newcommand{\Fz}{{\displaystyle \tF}^{(1)}} %  
\newcommand{\Fto}{{\displaystyle \tF}^{(t+1)}} %  

% memory
\newcommand{\M}{{\displaystyle \tM}} % 
\newcommand{\mem}{{\displaystyle \tM}} % 
\newcommand{\D}{\displaystyle \etD} % 
\newcommand{\Dz}{{\displaystyle \etD}^{(1)}} % 
\newcommand{\Dt}{{\displaystyle \etD}^{(t)}} % 
\newcommand{\Dto}{{\displaystyle \etD}^{(t+1)}} % 
\newcommand{\Dn}{{\displaystyle \etD}^{(n)}} % 
\newcommand{\Dno}{{\displaystyle \etD}^{(n+1)}} % 
\newcommand{\Mt}{{\displaystyle \tM}^{(t)}} % 
\newcommand{\Mtp}{{\displaystyle \tM}^{(t)'}} % 
\newcommand{\Mn}{{\displaystyle \tM}^{(n)}} % 
\newcommand{\Mno}{{\displaystyle \tM}^{(n+1)}} % 
\newcommand{\vMtk}{{\displaystyle \tM}^{(t)}_{\textrm{view}_k}}
\newcommand{\pMtk}{{\displaystyle \tM}^{(t)}_{\textrm{proj}_k}}
\newcommand{\viewtk}{{\displaystyle \etM}^{(t)}_{\textrm{view}_k}}
\newcommand{\viewn}{{\displaystyle \etM}^{(n)}_{\textrm{view}_n}}
\newcommand{\viewno}{{\displaystyle \etM}^{(n+1)}_{\textrm{view}_{n+1}}}
\newcommand{\viewnno}{{\displaystyle \etM}^{(n)}_{\textrm{view}_{n+1}}}
\newcommand{\viewfno}{{\displaystyle \etF}^{(n+1)}_{}}
\newcommand{\Mz}{{\displaystyle \tM}^{(1)}} % 
\newcommand{\Mto}{{\displaystyle \tM}^{(t+1)}} % 
\newcommand{\Mtwo}{{\displaystyle \tM}^{(t+1)}} % 

\newcommand{\Image}{\mathrm{I}}

\newcommand{\Mone}{{\displaystyle \tM}^{(1)}} % 
\newcommand{\len}{64} 
\newcommand{\model}{CoCoNets} 
\newcommand{\singlemodel}{CoCoNet} 
\newcommand{\votenet}{VoteNet} 
\newcommand{\ourvotenet}{\model-\textit{VoteNet}} 
\newcommand{\map}{\mathbf{M}} 
\newcommand{\basone}{3DVoxContrast} 
\newcommand{\bastwo}{3DPointContrast}
\newcommand{\basthree}{2.5DDenseObjectNets}
\newcommand\adam[1]{\textcolor{magenta}{#1}}
\newcommand\todo[1]{\textcolor{red}{#1}}
\newcommand\red[1]{\textcolor{red}{#1}}
\newcommand\gist[1]{\textcolor{cyan}{#1}}
\newcommand\ishita[1]{\textcolor{purple}{#1}}

\newcommand\topic[1]{\textcolor{cyan}{#1}}

\renewcommand{\arraystretch}{1.1}

%\makeatletter
%\@addtoreset{section}{part}
%\makeatother
%\titleformat{\part}[display]
%{\LARGE\bfseries\centering}{}{0pt}{}

%{\normalfont\LARGE\bfseries\centering}{}{0pt}{}

%\makeatletter
%\@addtoreset{section}{part}
%\makeatother

\newcommand{\viewpredcolwidth}{0.2\textwidth}

\begin{abstract}
This paper explores self-supervised learning of amodal 3D feature representations from RGB and RGB-D posed images and videos, agnostic to object and scene semantic content, and evaluates the resulting scene representations in the downstream tasks of visual correspondence, object tracking, and object detection. The model infers a latent 3D representation of the scene in the form of 3D feature points, where each continuous world 3D point is mapped to its corresponding feature vector. The model is trained for contrastive view prediction by rendering 3D feature  clouds in queried viewpoints and matching against the 3D feature point cloud predicted from the query view. Notably, the representation can be queried for any  3D location, even if it is not visible from the input view.  Our model brings together three powerful ideas of recent exciting research work: 3D  feature grids 
as a neural bottleneck for view prediction, implicit functions for handling resolution limitations of 3D  grids, and contrastive learning for unsupervised training of feature representations. We show the resulting 3D  visual feature representations  effectively scale across objects and scenes, imagine information occluded or missing from the input viewpoints, track objects over time, align semantically related objects in 3D, and improve 3D object detection. We outperform many existing state-of-the-art methods for 3D feature learning and view prediction, which are either limited by 3D grid spatial resolution,  do not attempt to build amodal 3D representations, or do not handle combinatorial scene variability due to their non-convolutional bottlenecks.
\end{abstract}

%\vspace{-1em}
\section{Introduction} \label{sec:intro}

%For visual perception models to generalize beyond the domain they are trained on, it is important for the models to learn or improve their feature representations without human annotations.  
%The current approach involves training on a pretext task which does not require human annotations, such as view/frame/context  prediction, with the hope of learning representations useful for downstream semantic tasks, such as object detection, tracking, manipulation and navigation  
%\cite{Eslami1204,commonsense,adam3d,lee2017unsupervised,DBLP:journals/corr/DoerschGE15,pathakCVPR16context,DBLP:journals/corr/ZhangIE16,DBLP:journals/corr/abs-1806-09594,oord2018representation,sermanet2018timecontrastive}.
%Many of these downstream tasks require reasoning about free space, scene and object occupancy, correspondence of objects and parts across time and space,  understanding camera motion, and invariance to cross-object occlusions \cite{weng2020ab3dmot,nagabandi2019deep}. 
Understanding the three-dimensional structure of  objects and  scenes  may be a key for success of machine perception and control  in object detection, tracking, manipulation and navigation.   
%these tasks, as well as generalization of perception across camera viewpoint and scene variations.  
Exciting recent works have explored learning representations of objects and scenes from multiview imagery and capture the three-dimensional scene structure implicitly or explicitly with  3D binary or feature grids \cite{DBLP:journals/corr/TulsianiZEM17,TulGupFouEfrMal17,sitzmann2018deepvoxels}, 3D point feature clouds \cite{Wiles_2020_CVPR},  implicit functions that map continuous world coordinates to  3D point occupancy  \cite{chibane2020implicit,genova2020local,sitzmann2020scene, mescheder2019occupancy, park2019deepsdf, chen2019learning}, as well as 1D or 2D feature maps \cite{Eslami1204}. 
These methods typically evaluate  the accuracy of the inferred 3D scene occupancy  \cite{chibane2020implicit,novotny2020canonical,DBLP:journals/corr/TulsianiZEM17, mescheder2019occupancy, park2019deepsdf, chen2019learning} and the fidelity of   image views rendered from the 3D representation \cite{le2020novel,mildenhall2020nerf,Eslami1204,Wiles_2020_CVPR,sitzmann2020scene}, as opposed to the suitability of representations for downstream semantic tasks. 
Methods that indeed focus on rendering photo-realistic images often give up on cross-scene generalization \cite{mildenhall2020nerf,riegler2020free}, or focus on single-object scenes \cite{sitzmann2020scene}. 
Methods that instead focus on learning semantically relevant scene representations are expected to generalize across scenes, and handle multi-object scenes. 
In the 2D image space, contrastive predictive coding  has shown to generate state-of-the-art visual features for correspondence and recognition \cite{oord2018representation,han2019video}, but does not encode  3D scene structure.  
In 3D voxel feature learning methods  \cite{adam3d,harley2020tracking}, convolutional latent 3D feature grids encode the 3D structure and a view contrastive objective learns semantically useful 3D representations, but  the  grid resolution limits the discriminability of the features learnt. 
Recent exciting works combine 3D voxel grids and implicit functions and learn to  predict 3D scene and object 3D occupancy from a single view with unlimited spatial resolution \cite{peng2020convolutional,popov2020corenet}.  
The model proposed in this work brings together these two powerful ideas: 3D  feature grids 
as a 3D-informed neural bottleneck for contrastive view prediction \cite{adam3d}, and implicit functions for handling the resolution limitations of 3D  grids \cite{peng2020convolutional}. % and contrastive predictive coding \cite{oord2018representation} for unsupervised training of 3D feature representations by predicting views.

We propose Continuous Contrastive 3D Networks (\model), a model that learns to map RGB-D images to infinite-resolution 3D scene feature representations by contrastively predicting views, in an object and scene agnostic way. 
% \ishita{Moreover, we will later show how \model{} can be extended to use just the RGB images for the task of view regression and point alignment.}
% that are useful for downstream tasks of recognition and detection.
% They lift RGB-D (2.5D) views to \todo{feature function grids and renders 3D point features to generate the corresponding 2D feature maps by concurrently training a 2D CNN in an end-to-end differentiable manner. }
Our model is trained to predict views of static scenes given 2.5D (color and depth; RGB-D) video streams as input, and is evaluated on its ability to detect and recognize objects in 3D. 
\model{} map the 2.5D input streams into 3D feature grids of the depicted scene. %At every frame, the architecture accounts for the motion of the camera, so that the internal 3D representation remains stable under egomotion. 
Given a target view and its viewpoint, 
the model first warps its inferred 3D feature map from the input view to a target view, then queries point features using their continuous coordinates,  and pulls these features closer to the point features extracted from the target view at the same 3D locations (Figure \ref{fig:model}). We use a contrastive loss to measure the matching error, and backpropagate gradients end-to-end to our differentiable modular architecture. At test time, our model forms plausible 3D completions of the scene given a \textit{single RGB-D image} as input: it learns to fill in information behind occlusions, and infer the 3D extents of objects. 
%Moreover, we show how \model{} can  be extended to use just the RGB images for the tasks of view prediction and point alignment. %\usepackage}

We demonstrate the advantages of combining 3D neural bottleneck, implicit functions and contrastive learning for 3D representation learning 
by comparing our model against state-of-the-art self-supervised models, such as i) contrastive learning for pointclouds \cite{xie2020pointcontrast}, which shares a similar loss but not the amodal predictive ability of our model, ii) contrastive neural mapping \cite{harley2020tracking}, which can amodally inpaint a 3D discrete feature grid but suffers from limited spatial grid resolution, and iii) dense  ObjectNets \cite{pmlr-v87-florence18a}, which self-learns  2D (instead of 3D) feature representations with a triangulation-driven supervision similar to (i).  
Our experimental results can be summarized as follows:
\textbf{(1)} 3D object tracking and re-identification (Figure \ref{fig:tracking_quali}): We show that   scene representations learnt by \model{} can detect objects in 3D across large frame gaps better than the baselines  \cite{xie2020pointcontrast,harley2020tracking,pmlr-v87-florence18a}.
%outperform the SOTA method in tracking 3D objects. 
\textbf{(2)} Supervised 3D object detection: Using the learnt 3D point features as initialization  boosts the performance of the state-of-the-art Deep Hough Voting detector of \cite{qi2019deep}. \textbf{(3)} 3D cross-view and  cross-scene object 3D alignment: We show that the learnt 3D feature representations can infer 6DoF alignment between the same object in different viewpoints, and across different objects of the same category, better than  \cite{xie2020pointcontrast,harley2020tracking,pmlr-v87-florence18a}. 
 %, and show superior performance %With these comparisons we demonstrate the advantages of combining 3D neural bottleneck, implicit functions and contrastive learning for 3D representation learning. 
We further show that our model can predict image views (with or without depth as input) and 3D occupancies that outperform or are on par with the state-of-the-art view and occupancy prediction models \cite{popov2020corenet,commonsense,Eslami1204}.

In summary, the main contribution of this paper is a model that learns  infinite-resolution 3D scene representations from  RGB-D posed images,  useful for tracking and corresponding objects in 3D, pre-training  3D object detectors, and predicting views and 3D occupancies. We set a new state-of-the-art  in self-supervision of 3D feature representations. %Our data and code will be made publicly available.%Our model further predicts RGB image views and 3D occupancies and can generalize to truly novel scenes. %To the best of our knowledge, our work is the first to learn amodal 3D representations are the first to extract high-quality  3D features.
\section{Related work} \label{sec:related}
% \thispagestyle{empty}
%We present the body of related literature with respect to the goal. Each approach is designed to achieve: 3D occupancy prediction, RGB image generation, and visual feature representation learning. 
%The inductive biases are shared across 
%Methods in the above three categories have explored 3D inductive architectural biases, as we detail below.

\paragraph{Learning to 3D reconstruct objects and scenes}
%\topic{goal: 3D reconstruction extraction}
Learning to infer  3D reconstructions of objects and scenes from  single images or videos  has been the goal of recent deep geometrical methods, that have explored a variety of explicit 3D representations, such as 3D point clouds \cite{DBLP:journals/corr/NovotnyLV17a,lin2017learning}, 3D binary voxel occupancies \cite{DBLP:journals/corr/TulsianiZEM17,yan2017perspective}, or 3D meshes \cite{DBLP:journals/corr/abs-1711-07566,DBLP:journals/corr/abs-1906-02739}. Since detailed 3D supervision is only possible in large scale  in simulation, many  approaches  attempt supervision from  RGB or depth map prediction through differentiable rendering of the inferred 3D reconstruction  \cite{DBLP:journals/corr/abs-1711-07566,DBLP:journals/corr/abs-1906-02739}. To handle limitations of spatial resolution of 3D voxel grids,  
recent approaches represent 3D object occupancy with implicit functions parameterized by deep neural networks  trained to map continuous world 3D coordinates to the corresponding point 3D occupancy values \cite{chibane2020implicit, mescheder2019occupancy, park2019deepsdf, chen2019learning}. While these methods train one 3D shape implicit function per object, by assigning a 1D latent embedding to each object, some works \cite{popov2020corenet,peng2020convolutional} train 3D grids of  functions where each function takes care of estimating the occupancy of points in the vicinity of the corresponding 3D voxel centroid. These latter methods enjoy the generalization of 3D convolutions and can scale to multi-object scenes, while implicit functions parametrized by non-convolutional, fully connected networks are mostly limited to single object scenes \cite{chibane2020implicit}. 
Our approach also employs 3D grids of functions which predict feature embeddings for the corresponding continuous 3D world coordinates as opposed to merely occupancy.
%\todo{do they attempt reconstuction or densification only? We need to add 2 lucey papers, anything else here?}

%\topic{goal: awesome image renders}
\paragraph{Neural image synthesis with 3D inductive biases}
Deep image generative networks have shown compelling results in generating photorealistic images that match the image statistics of the unlabelled  image collections they are trained on \cite{NIPS2014_5423}. They are based on variational autoencoders \cite{kingma2013auto}, generative adversarial networks \cite{brock2019large}, generative flows \cite{kingma2018glow}, or autoregressive image pixel generators \cite{van2016conditional}. However, these 2D generative models  learn to parameterize the manifold of 2D natural images, and struggle to generate images that are multi-view consistent, since the underlying 3D scene structure cannot be exploited. 
% \thoughts{Recent works \cite{nguyen2019hologan, schwarz2020graf} show that they can perform 3D aware image synthesis just by training on unposed 2D images. These generative modelling approaches disentangle the scene properties from viewpoint, allowing the content to be preserved across viewpoints. These approaches, however, are restricted to single object scenes.}
Generative models trained from multi-view data can render arbitrary views of an input scene \cite{tewari2020state, LSM,sitzmann2018deepvoxels,DBLP:journals/corr/TatarchenkoDB15,DBLP:journals/corr/abs-1905-05172}. Such multi-view generative models often restrict themselves to single-object scenes \cite{sitzmann2020scene,nguyen2019hologan,schwarz2020graf} or to a single complex scene without aiming at cross-scene generalization \cite{mildenhall2020nerf,riegler2020free}, with the goal of generating high fidelity photorealistic images, replacing hand-engineered graphics engines. Their architectures incorporate many inductive biases of graphics engines, such as 3D-to-2D rendering modules \cite{tewari2020state} and  explicit feature transformations to handle  viewpoint changes \cite{commonsense}. Their lack of cross-scene generalization or their limitation to single object scenes makes it hard to adopt their inferred feature representations for visual recognition. 
% We also show that generating photorealistic images and learning useful visual representations are not compatible learning objectives. %, and in fact we train two separate network, one predicting images, and the other predicting constrastive feature representations. 
% , since in that case we would hope view prediction to precede, not to succeed object detection. Our model uses view prediction to help object detection, not the inverse. 

\paragraph{Learning visual feature representations  by self-supervised view prediction}

Recent methods learn neural scene representations by predicting views of a scene under known egomotion \cite{Eslami1204,sitzmann2020scene,adam3d,sitzmann2018deepvoxels}. View prediction, as opposed to the related and very effective objective of feature learning via triangulation \cite{xie2020pointcontrast}, results in explicitly or implicitly \textit{amodal} representations, i.e., representations that can predict  information missing from the input observations \cite{doi:10.1177/2041669518788887}, as opposed to simply featurizing the visible image pixels \cite{pmlr-v87-florence18a}. 
Different view prediction methods for learning representations vary with respect to the amount of their reasoning  regarding geometry and  the underlying 3D structure of the scene \cite{tobin2019geometryaware}.  
The generative query network (GQN) of Eslami \etal \cite{Eslami1204} showed that it can predict alternative views of toy simulated scenes without explicit 3D structure inference, and demonstrated the usefulness of the inferred representations as pre-training for reinforcement behaviour learning tasks. 
Geometry-aware recurrent networks of Tung \etal \cite{commonsense} use a latent 3D feature map in their bottleneck for view prediction and demonstrate superior generalization, granted from the 3D convolutional inductive bias. Harley \etal \cite{adam3d} uses a similar 3D latent feature grid but optimizes for contrastive prediction as opposed to RGB regression, and demonstrate its usefulness for 3D object tracking and 3D moving object segmentation. %Contrastive prediction introduced in Oord \etal's work  \cite{oord2018representation} optimizes an objective that preserves the mutual information between ``top-down'' contextual features inferred from input observations, and ``bottom-up'' features inferred from future observations, and obtained state-of-the-art results and applied it in various domains, such as predicting speech, text, and image patches, with success. 
Our work optimizes a contrastive view prediction objective similar to \cite{adam3d} but uses a 3D grid of implicit functions as its latent bottleneck. We empirically show that the emergent 3D feature representations are more accurate in 3D object tracking and visual correspondence than the features obtained from existing state-of-the-art 3D feature learning methods.
% \todo{we should be more specific here at the end}

%The latter is the closest work to ours, and show that the addition of implicit function in the 3D latent bottleneck has a very large impact on the performance of learnt features, as validated in our experimental section. 

\section{Continuous Contrastive  3D Networks (\model) for Learning Amodal Visual Representations}

\label{sec:model}

\begin{figure}
  \begin{center}
    \includegraphics[width=0.5\textwidth]{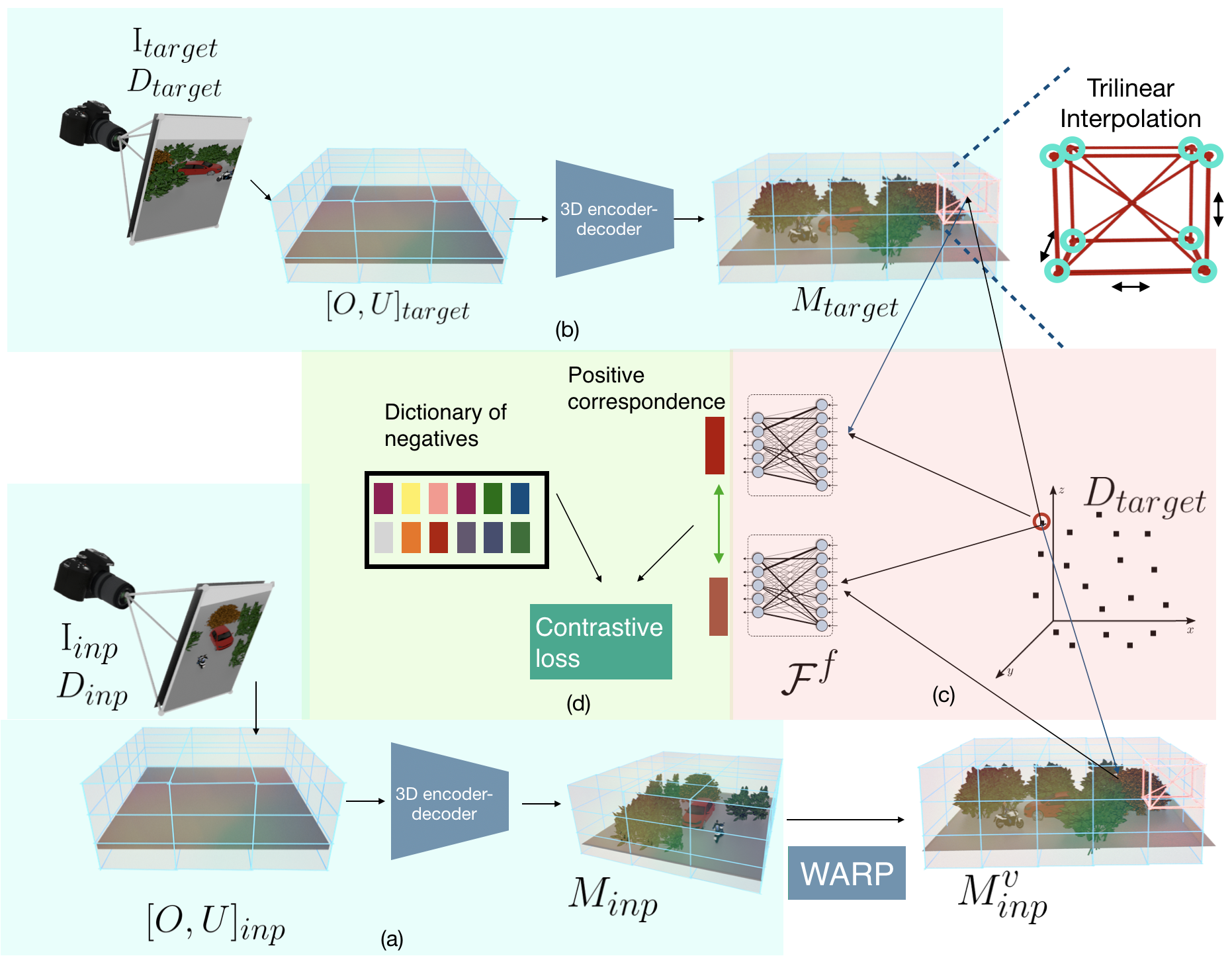}
  \end{center}
 \caption{\textbf{Continuous Convolutional Contrastive  3D Networks} (\model)  are trained to lift 2.5D images to 3D feature function grids of the scene by optimizing  for view-contrastive prediction. 
(a) In the top-down path, the model encodes RGB-D  images into a 3D feature map $\map \in  \mathbb{R}^{w \times h \times d \times c}$, and uses explicit  3D feature transformations (translation and 3D rotation) to account for changes of viewpoint between the input and target views.  
 (b) In the bottom-up path, we encode the RGB-D of the target viewpoint into a 3D feature cloud.
 (c) Given  continuous 3D world coordinates $(X,Y,Z)$ and its embedded code $v_{(X,Y,Z)}$ inferred via trilinear interpolation, a fully connected network maps the coordinates and the embedded code, to the feature vector of the 3D point at location $(X,Y,Z)$.
 %which is projected to the corresponding pixel location $x,y$.
 %encoding of the othat represent an offc of points in its vicinity, it predicts the corresponding 3D point feature vectors  It then samples 3D point features by sampling continuous 3D world point coordinates and concatenating those with the feature vector  projects this volume into a featuremap of the target viewpoint. 
 (d) Metric learning losses in 3D tie the two point cloud representations together.}
\label{fig:model}
\end{figure}

We consider a mobile agent that can move about the scene and has access to its egomotion. The agent has a color camera with known intrinsics, and a depth sensor registered to the camera's coordinate frame. 
We use groundtruth depth provided by the simulation environment, and we will show in Sec.~\ref{sec:exp} that the learned models generalize to the real world, where (sparser) depth is provided by a LiDAR unit. %\todo{We further show our model can operate without depth at test time} in Sec.~\ref{sec:exp} by encoding RGB images directly to 3D  function grids. 

\model\space learn 3D  visual feature representations by collecting posed images in static scenes and doing contrastive view prediction. 
We describe the  architecture in Sec.~\ref{sec:model1}. %and the view-contrastive prediction objectives in Sec.~\ref{sec:model2}.  
We then evaluate the correspondability of the resulting 3D feature representations in  3D object re-identification and tracking in dynamic scenes 
(Section \ref{sec:tracking_exp}), as pre-training for 3D object detection  (Section \ref{sec:objdet_exp}), and cross-object semantic visual correspondence  (Section \ref{sec:corres_exp}). 

%and our unsupervised 3D object tracking and cross instance correspondence methods method in Sec.~\ref{sec:motionseg}.

\subsection{Continuous Contrastive  3D Networks (\model)}
\label{sec:model1}

Our model's architecture is illustrated in Figure~\ref{fig:model}. It is a  neural network with a three-dimensional neural bottleneck $\map \in \mathbb{R}^{w \times h \times d \times c}$, which has three spatial dimensions (width $w$, height $h$, and depth $d$) and a feature dimension ($c$ channels per grid location). 

The latent state aims to capture an informative and geometrically-consistent 3D deep feature map of the world space. 
Therefore, the spatial extent corresponds to a large cuboid of world space, defined with respect to the camera's position at the first timestep.  
Each voxel in the 3D feature map $\map$ corresponds to a cuboid in the 3D scene depicted in the RGB-D image. 

To be able to generate features at infinite spatial resolution, i.e., to featurize continuous 3D physical points within a voxel, we use implicit function parametrization. 
%\textbf{3D point feature  generation \enskip} peng2020convolutional
Our model's architecture of interpolating within a voxel grid resembles that of Peng \etal \cite{peng2020convolutional}. Let $(X,Y,Z)$ denote the continuous world coordinate of a 3D point whose feature we wish to infer. First, we trilinearly interpolate the feature grid to obtain a  a $c$-dimensional feature vector at point $(X,Y,Z)$.  Denoting this trilinearly interpolated feature vector as $p$ for input point $(X,Y,Z)$, we further obtain a 3d location conditioned feature vector using $\phi(p, x)$, where $\phi$ is a small fully-connected network. Finally we obtain our 32-dimensional embedding vector using $f_\theta(p,\phi(p, x))$, where $f_\theta$ represents a multi-block fully connected ResNet. Further details
about the network architecture can be found in the supplementary.

% Let $(X,Y,Z)$ denote the continuous world coordinate of a 3D point whose feature we wish to infer. First, we compute the interpolated version of the feature vector at $(X,Y,Z)$ of the  voxel that contains the point in the 3D feature map, and  $(dX,dY,dZ)=(X,Y,Z)-(i,j,k)$ is the \textit{offset} from that containing voxel's center. We encode the offset into a $c$-dimensional vector $v_{(dX,dY,dZ)}$ using a (trainable) linear layer. 
% %with continuous coordinates , we first map it to its coordinates $(X,Y,Z)$ in the 3D voxel grid. The voxel identity $(i,j,k)$ in which it lies is then just the integer part of $(X,Y,Z)$. Then,  we compute the offset $(dX,dY,dZ)=(X,Y,Z)-(i,j,k)$, and encode it into a $c$-dimensional vector $v_{(dX,dY,dZ)}$ using a linear layer.
% The offset code $v_{(dX,dY,dZ)}$, the feature $\map_{i,j,k}$ at location $(i,j,k)$ in the voxel grid, and the interpolated feature vector at location $(X,Y,Z)$ then pass through a five-layer, fully-connected ResNet  \cite{he2016deep, peng2020convolutional} to predict the final feature vector of the 3D point at location $(X,Y,Z)$. 

A similar fully-connected ResNet parametrization provides a point's binary  occupancy $o_{(X,Y,Z)}$, and its RGB color value $c_{(X,Y,Z)}$. The three ResNets that predict features, occupancy and RGB values of 3D points do not share weights. During training, we use the point cloud of the target viewpoint to query our model for features, occupancies and colors in the corresponding visible  3D point locations and propagate gradients in an end-to-end manner to the same feature voxel 3D map $\map$. We denote the operation of obtaining point features, occupancies and colors by querying the feature map at continuous locations by $\F^f\left(\map,(X,Y,Z)\right),\F^o\left(\map,(X,Y,Z)\right),$ and $\F^c\left(\map,(X,Y,Z)\right)$.
At test time, we query the model in both visible and non visible 3D point locations, to obtain an amodal completed 3D point feature cloud. %have a predefined set of 200K

% occupancies and % colors we use a predefined set of 3D point locations on which we query our model, and project them using visibility aware raycasting to the target viewpoints. \todo{We query the memory state of our model using different locations of 3D points, In }

%We refer to the latent state $\Mt$ as the model's \textit{imagination} to emphasize that 

\model{} is made up of differentiable modules that go back and forth between 3D feature  space and 2D image space. 
It can take as input a variable number of RGB-D images both at training time and test time. For simplicity, in our experiments we encode only a single view input at both train and test time. More details on each of the following modules are included in the supplementary file. 

\begin{figure*}[h!]
  \centering
  \includegraphics[width=0.9\textwidth]{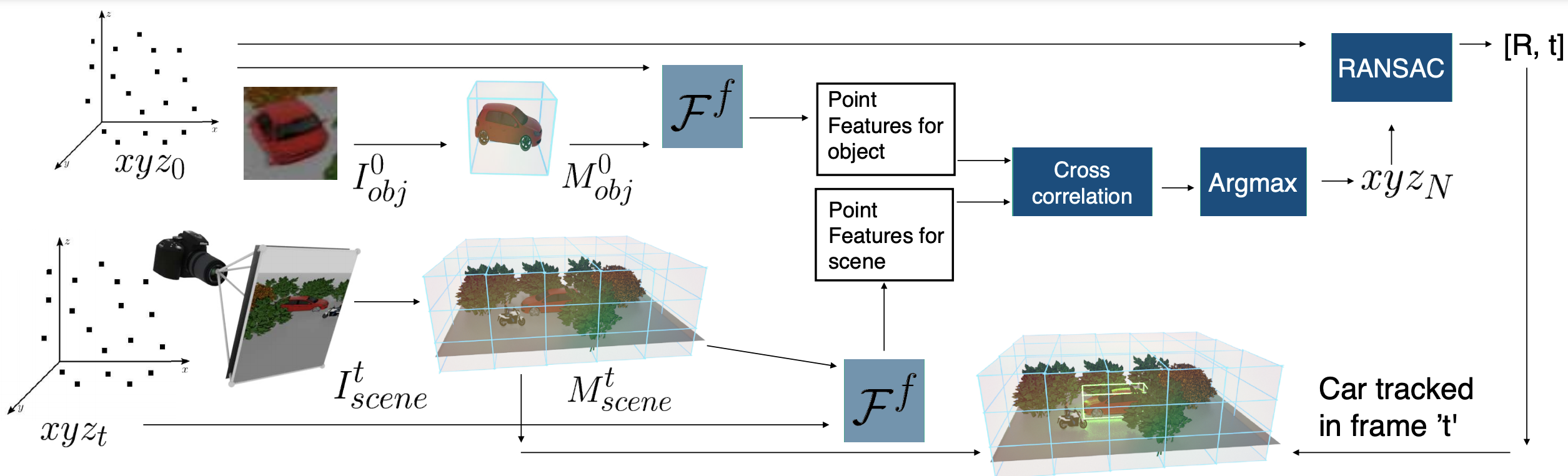}
  \caption{  \textbf{3D object tracking using \model.} Given the cropped RGB-D image $I_{obj}$ of the object to track at $t = 0$, our model infers the 3D object feature map $M_{obj}$, and queries it using $xyz_0$, the point cloud of the object, to obtain object point features. Similarly, it obtains the point features of the entire scene at timestep $t$. Finally, it does cross correlation between these features to get $xyz_N$, where each $i^{th}$ point in $xyz_N$ is the point from the scene whose feature matched best with the feature for $i^{th}$ point in $xyz_0$. We then apply RANSAC on $xyz_0$ and $xyz_N$ to obtain the location of the car at timestep $t$.
  }
  \label{fig:trackingdetails}
\end{figure*}

\textbf{2D-to-3D unprojection \enskip} (Figure \ref{fig:model} (a, b))
This module converts the input RGB image $\Image \in \mathbb{R}^{w \times h  \times 3}$ and  depth map $\D \in \mathbb{R}^{n \times 3}$ where, $n$ is the size of the pointcloud) into 3D tensors using available camera intrinsics.
The RGB is ``unprojected'' into a 3D tensor $\mathbf{U} \in \mathbb{R}^{w \times h \times d \times 3}$ by filling each 3D grid location with the RGB value of its corresponding subpixel.
%Specifically, for each cell in the imagination grid, indexed by the coordinate $[i,j,k]$, we compute the floating-point 2D pixel location $[u,v]^T = K R [i,j,k]^T$ that it projects to from the current camera viewpoint, where $R$ is the similarity transform that converts memory coordinates to camera coordinates and $K$ is the camera intrinsics (transforming camera coordinates to pixel coordinates). 
The pointcloud is converted to a 3D occupancy grid $\mathbf{O} \in \mathbb{R}^{w \times h \times d \times 1}$, by assigning each voxel a value of 1 or 0, depending on whether or not a point lands in the voxel. 
%\todo{do we ever use only RGB.If a depth map is not available, we replicate the RGB pixel value all along the unprojected ray. }
% 3D tensor $[\Ut, \Ot]\in \mathbb{R}^{w \times h \times d \times 4}$. 
%We assume a pinhole camera model \citep{hartley2003multiple},
%We fill $\Ut_{i,j,k}$ with the bilinearly interpolated pixel value  $\It_{u,v}$. 
%We transform our depth map $\D_t$ in a similar way and obtain  a binary occupancy grid $\Ot \in \mathbb{R}^{w \times h \times d \times 1}$, by assigning each voxel a value of 1 or 0, depending on whether or not a point lands in the voxel. We concatenate this to the unprojected RGB, making the tensor $[\Ut, \Ot] \in \mathbb{R}^{w \times h \times d \times 4}$.
%\textbf{3D encoder-decoder \enskip} 
%This module converts the raw 3D input tensors $[\Ut, \Ot]\in \mathbb{R}^{w \times h \times d \times 4}$ into a 
We then convert the concatenation of these tensors into a 3D feature tensor $\map \in \mathbb{R}^{w \times h \times d \times c}$, via a 3D convolutional encoder-decoder network with skip connections. We $L_2$-normalize the feature in each grid cell. %, producing a 3D feature tensor for the timestep, denoted 

\textbf{3D-to-2D projection \enskip}
This module first warps the 3D feature map $\map$ to align it to the target viewpoint, yielding $\map^{\mV}$.
It then generates  a 3D feature point cloud by querying the model at the 3D point locations visible from the target viewpoint, provided by the target depth map $\D_{target}$.
It also generates a 2D feature map  by 
%We also project the corresponding features and obtain
%and a 2D feature maps as follows 
%, and then map it to a 2D feature map $\viewtk$ with a 2-block 2D ResNet \cite{he2016deep}.
%It then 
computing for each visible 3D point the corresponding 2D pixel location  using the camera intrinsics, $(x,y)=\mathrm{f}\frac{X}{Z},\mathrm{f}\frac{Y}{Z}$, and copying the 3D point feature into that location in the 2D feature map. 
%feature vectors of  the points to corresponding 2D pixel locations using the camera intrinsics, that is, camera focal length $\mathrm{focal}$, camera center $(u_x,u_y)$, and a scaling factor $\alpha$ \todo{we need the camera projection equations here}:
Note that each 3D feature point is mapped independently to 2D, in contrast to neural renderers that convolutionally map feature maps to image views \cite{sitzmann2018deepvoxels, Eslami1204}. This independent point-by-point rendering is closer to the spirit of graphics operations \cite{sitzmann2020scene}. %and rendering is more well defined 
%in contrast to learning based neural r

%state $\Mt$ into a 2D feature map of a desired viewpoint $\Vk$. 

%We first warp the 3D feature map $\Mt$ into a view-aligned version $\vMtk$, then map it to a 2D feature map $\viewtk$ with a 2-block 2D ResNet \cite{DBLP:journals/corr/HeZRS15}. %We denote the final output of this 3D-to-2D process as .

%% \subsection{View-contrastive 3D predictive learning}
%\subsection{Contrastive Predictive Training} %\label{sec:model2}

%\todo{we need to write about Phil's sampling}

%We measure 
\textbf{3D contrastive learning of point features \enskip}
Given a pair of input RGB images $(\Image_{inp}, \Image_{target})$, depth maps $(\D_{inp}, \D_{target})$, and the camera pose change between input and target views $\mV$, 
%, we train our model to predict feature abstractions of an unseen input $(\Ino, \Dno, \Vno)$, as shown in Figure~\ref{fig:model}-left. 
we consider two types of representations for the target view:
\begin{itemize}
    \item a top-down one, $\map_{inp}^{\mV}$ (Figure \ref{fig:model} (a)) by encoding the input RGB-D image $(\Image_{inp},\D_{inp})$ and orienting the feature map to the target viewpoint, and predicting the features for the target 3D points in $\D_{target}$ by querying the functions in $\map_{inp}^{\mV}$, obtaining the feature cloud $\{   \left( X,Y,Z,\F(\map_{inp}^{\mV}, (X,Y,Z)) \right) \}$    for  $(X,Y,Z) \in \D_{target}$ (Figure \ref{fig:model} (c)). 
    
    \item a bottom-up one, $\map_{target}$ (Figure \ref{fig:model} (b)) by simply encoding the target RGB-D image   $\Image_{target},\D_{target}$ and predicting the features for the target 3D points in $\D_{target}$, obtaining the feature cloud
    $\{  \left(X,Y,Z,\F(\map_{target}, (X,Y,Z)) \right) \}$    for  $(X,Y,Z) \in \D_{target}$ (Figure \ref{fig:model} (c)). 
    %by querying the functions in $\M_{target}$. 
\end{itemize}
We use a contrastive InfoNCE loss \cite{oord2018representation} (Figure \ref{fig:model} (d)) to pull corresponding top-down and bottom-up point features close together in embedding space and push non-corresponding ones farther away:
%our metric learning loss evaluates the matching of the context-based and bottom-up representations: 
% \begin{small}
% \begin{align}
%     \loss& = 
%     \sum_{i,j,k,m,n,o}
%     \max ( \mathcal{Y}\three_{ijk,mno}  (\| \F^f(\map_{target}, (i,j,k)) -\\ \nonumber &\F^f(\map_{inp}^{\mV}, (m,n,o))\|_2 - \alpha), 0  ),\\ \nonumber
% \end{align}
% \end{small}
% \begin{small}
\begin{align}
    \loss& = 
    -\log   \frac{\exp(\frac{q \cdot k_+}{\tau})}{\sum_{i=0}^{K}\exp{\frac{q \cdot k_i}{\tau}}} ,
\end{align}
where $\tau$ is a temperature hyper-parameter and the sum in the denominator is over one positive and $K$ negative samples. In the numerator, $q$ represents $\F^f(\map_{inp}, (i,j,k))$, and $k_+$ is its corresponding positive sample $\F^f(\map_{target}, (m,n,o))$. The dot product computes a similarity. % is computed using a dot product. 
% \end{small}
% where $\tau$ is the margin size, and $\mathcal{Y}$ is $1$ at continuous 3D coordinates
% for positive pairs and $-1$ for negatives.
% \sout{where $\ctx$ corresponds to $\bot$, and $-1$ everywhere else.}
%We can get an additional self-supervision training signal from the fact that features corresponding to the same 3D physical point, when sampled from any two 3D feature grids $\Fi$ and $\Fj$, should be closer to each as compared to any randomly sampled 3D points. We achieve this by 
In practice, we randomly sample corresponding points from the top-down or bottom-up feature clouds at each training iteration.
%, calculating the features for these points using $\Fi$ and $\Fj$, and treating them as positive pairs for 3D contrastive learning. 
We also maintain a large pool of negative pairs in a dictionary, using the approach proposed by He \etal
\cite{he2020momentum}.

%The losses ask tensors depicting the same scene, but acquired from different viewpoints, to contain the same features. 

%The performance of a metric learning loss depends heavily on the sampling strategy used \cite{schroff2015facenet,DBLP:journals/corr/SongXJS15,sohn2016improved}. We use the distance-weighted sampling strategy proposed by \cite{wu2017sampling} which uniformly samples ``easy'' and ``hard'' negatives; % \citep{wu2017sampling,lehnensphere};
%we find this outperforms both random sampling and semi-hard sampling \cite{schroff2015facenet}. 

Our  metric learning loss, if applied only on \textbf{3D points visible from both input and target views}, coincides with the point contrastive metric learning of Xie \etal \cite{xie2020pointcontrast}, a state-of-the-art 3D point feature learning method. \model{}  can handle input and target views that actually have few or no points in common. We compare against Xie \etal's  state-of-the-art 3D feature learning model in our experiments, and demonstrate the importance of amodal completion for feature learning.

%\textbf{3D contrastive learning for point features \enskip}
%We can get an additional self-supervision training signal from the fact that features corresponding to the same 3D physical point, when sampled from any two 3D feature grids $\Fi$ and $\Fj$, should be closer to each as compared to any randomly sampled 3D points. We achieve this by randomly sampling some points from any single view point cloud, calculating the features for these points using $\Fi$ and $\Fj$, and treating them as positive pairs for 3D contrastive learning. We also maintain a large pool of negatives pairs in a dictionary using the approach proposed by He \etal \cite{he2019momentum}.
\vspace{-0.15in}
\paragraph{Occupancy prediction and RGB view regression}
We train a separate \singlemodel{} model to predict occupancy and RGB  in novel viewpoints, which we denote \model-\textit{OccRGB}. We use a standard binary cross entropy loss for the occupancy, and a regresion loss for RGB. The model predicts point clouds in target viewpoints and infers their occupancy and color from the implicit function grids $\F^o,\F^c$. %Let \model$^{RGB-O}$ denote this model. 
Specifically, to predict a  target RGB view and its occupancy given a source RGB-D image, after encoding and orienting the 3D feature map $\map$ to the target viewpoint, 
we predict the point occupancies of the target view,  and  predict the point RGB colors for the occupied points. These  are then projected to the image space in the target view.

%We did not find improved feature representations by 
Jointly training RGB and occupancy prediction  with  contrastive view prediction did not improve discriminability of our feature representations. 
% Indeed, specularities and shadows often make the color of the same 3D physical point to differ across viewpoints. %Let \model{} ^{RGB+occ}
%the RGB and occupancy prediction model as 
The novelty of our paper is in using voxel-implicit  architectures for amodal contrastive 3D feature learning as opposed to 3D occupancy prediction \cite{popov2020corenet,peng2020convolutional}, or in addition to it. %---or in addition to---3D occupancy prediction,  already proposed in \cite{popov2020corenet,peng2020convolutional}.

%better than depth-augment GQN 
\paragraph{RGB view regression without depth input} Additionally, we also make use of NeRF's volumetric renderer \cite{mildenhall2020nerf}, with features extracted from $\map_{input}$ as priors, to render novel views using just a single RGB image (no depth information) as input. We call this \model$^{NoDepthRGB}$ and explain the setup in the supplementary file. Qualitative results from this model can be seen in Figure~\ref{fig:rgbnerf}.

\section{Experiments}\label{sec:exp}
% \vspace{-0.1in}
% \thispagestyle{empty}
Useful visual feature representations are expected to  correspond visual entities across variations in pose and appearance, as well as across  intra-category variability. We evaluate the correspondability of our 3D scene representations in 3D object tracking and object re-identification,  3D pose estimation and cross-object correspondence. We further evaluate our features as pre-training for 3D object detector for the state-of-the-art method of \cite{qi2019deep}.
Our experiments aim to answer the following questions:
\begin{enumerate}
\item Is amodal completion important for learning  scene  representations useful for  visual correspondence, across time and across scenes? To this end, we compare  3D feature representations learned by \model{} against  contrastive point clouds \cite{xie2020pointcontrast} that do not consider amodal completion but share a similar contrastive objective for 3D points visible in both scenes.
%the visible points where positive matches are  triangulation. 
  
\item Is our 3D function parameterization important for learning discriminative visual representations? To this end, % for  visual correspondence across time and across scenes?
we compare with  3D contrastive neural mapping of Harley \etal \cite{adam3d} that does not consider continuous functions, but rather uses discrete 3D feature maps as the neural bottleneck. %These comparisons 
%learning on point clouds  or amodal discrete 3D feature contrastive mapping of \cite{}

\item Does the performance of 3D object detectors in pointclouds  improve when using \model{} as initialization of the point features, and by how much? To this end, we compare with \votenet{}\cite{qi2019deep}, a model that uses PointNet++ \cite{qi2017pointnet++} as their feature extractor and deep Hough voting for 3D object detection in point clouds.

\item Is the continous 3D convolutional bottleneck of \model{} useful for view prediction?  To this end, we compare image views rendered from our model in novel scenes with images from GQN \cite{Eslami1204}, which does not consider 3D neural bottleneck, and GRNN \cite{commonsense}, which has a discrete 3D convolutional bottleneck.
    
\end{enumerate}

We train and test \model{} in the simulated  datasets of CARLA \cite{Dosovitskiy17},  ShapeNet \cite{shapenet2015} and test them further---without training---in the real-world  KITTI dataset \cite{Geiger2013IJRR}. 
CARLA  is an open-source photorealistic simulator of urban driving scenes, which permits moving the camera to any desired viewpoint in the scene. ShapeNet  is a large scale 3D object repository which again permits camera placement at will. KITTI is an urban driving scene dataset collected from a camera mounted on a car. More information on datasets is available in the supplementary file.
%We collect evideo sequences from CARLA's Town1 as the training set, and videos from Town2 as the test set.
%Specifically, 

% \subsection{Self-supervised 3D object tracking}
\subsection{Self-supervised 3D object tracking and re-indentification}
\label{sec:tracking_exp}
\paragraph{Setup}
We evaluate the 3D feature representations of \model{} in their ability to track a 3D object over time, as well as across large frame gaps, a task known as object re-identification. 
Given the 3D bounding box of an object in the first frame, we are interested in tracking the object in subsequent frames. We first infer 3D features for object points visible in the first frame. For subsequent timesteps, we predict the amodal  \textit{complete} 3D feature  cloud of the scene from a single view by querying our model in a set of visible and invisible 3D point locations. We found this uniform random sampling of query point set strategy to suffice for our tasks, though more elaborate strategies that focus on querying points close to object surfaces can be used, such as Marching Cubes \cite{10.1145/37402.37422}, which has been used by previous 3D implicit shape function models for learning object occupancy to obtain a watertight mesh \cite{genova2020local,peng2020convolutional,popov2020corenet}. Given the object feature point cloud and the amodal scene feature cloud,  we obtain a rigid body transformation (3D rotation and translation) by solving the orthogonal Procrustes problem \cite{schonemann1966generalized}, and use RANSAC to find the transformation that satisfies most inliers as shown in Figure \ref{fig:trackingdetails}, 
where inlier point matches are obtained by thresholding inner products at 0 between the object 3D point features and 3D point features of the target scene. We find the object 3D box in the target scene by  warping the box $b_0$ using the obtained transformation. Note that we do not restrict our model to search locally only for a matching object, but search over the whole scene which makes the task harder. Such object matching across frame gaps is  important for re-detecting an object after occlusions. It is also a good test for the discriminability of the features, since they are not assisted by temporal continuity or local search regions. 
\paragraph{Evaluation and comparison with baselines}
% We describe the evaluation setup for 3D object tracking and identification in Sec. \ref{sec:model2}.
%We evaluate \model{} in its ability to track a vehicle in 3D. Given the 3D box of an object at zero$^{th}$ frame, we are interested in tracking the object in subsequent frames. Tracking is a good task to evaluate how discriminative the features are. More discriminative and stable the features over semantics and time respectively, better the tracking will be. Semantically discriminative features ensure accurate matching of corresponding points, and stability of features across timesteps ensure we will be able to generate these point correspondence over multiple timesteps. For tracking task, 
We compare \model{} against the following baselines (we use names indicative to the differences and characteristics):
i) \basone{} of Harley \etal \cite{harley2020tracking} ,
ii) \bastwo{} of Xie \etal \cite{xie2020pointcontrast} 
%and some more baselines that \cite{harley2020tracking}  proposes in their paper\todo{explain each baseline?}.
iii) \basthree{} of Florence \etal \cite{pmlr-v87-florence18a}, which uses triangulation supervision to train 2D deep feature maps, then lifts them to 3D point feature clouds using the available depth map.

We  follow the experimental setup of Harley \etal \cite{harley2020tracking}, which uses video sequences of 10 timesteps. 
For \basone{}, we  first use multiview RGB-D data with known egomotion to train a model  using 3D contrastive losses. Upon training, \basone{} maps a single RGB-D image to a discrete 3D feature map.  Then, \basone{} uses this model for tracking by solving orthogonal procrustes problem with RANSAC between voxel features as opposed to point features. 
For \bastwo{}, upon training the model, we compute RANSAC on the cross-correlation scores between point features from the visible 3D points on the object at frame 0 and the \textit{visible} point cloud at each subsequent frame. Our method also considers amodal completed point cloud at each frame.
For \basthree{}, upon training the model, we apply RANSAC much the same way as in the \bastwo{} model.

%by creating feature grid $M_i$ at every timestep $i$. Then, they compute  cross-correlation between $M_i$ and vehicle features cropped from $M_0$ using the given box, apply RANSAC to find a rigid body transformation that satisfies most of the cross-correlations, and warp the box using this transformation to the current frame. 

%Instead of using the voxel features for computing cross correlations, we instead use \model{} to get point features for points in the point cloud in an implicit manner. We first pre-train \model{} using 3D contrastive losses (Section \ref{sec:ml3d}). Then, when tracking, we use only the features from object points visible in the zeroth frame. For subsequent timesteps/frames, we first predict the complete point cloud of the scene from single view \todo{explain in model section and refer here}, calculate point features for this completed point cloud, and then use these and object point features for computing cross correlation. 

\begin{figure}[h!]
  \centering
  \includegraphics[width=0.48\textwidth]{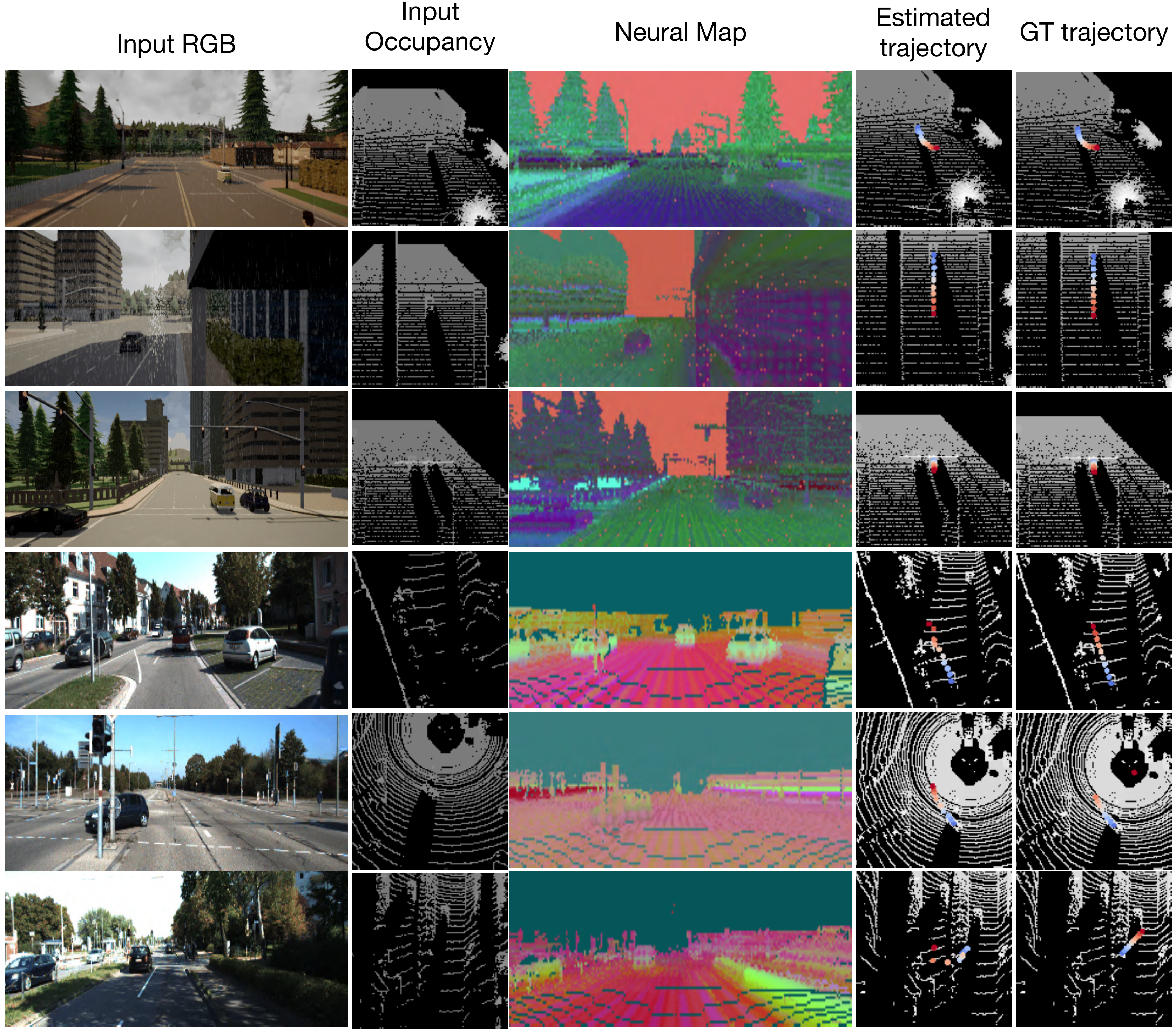}
  \caption{ \textbf{Self-supervised 3D object tracking}. In the 1st and 2nd column we visualize the RGB and depth from the first frame, which is given as input to our model, along with a 3D box specifying the object to track. In the 3rd column we visualize our inferred point features by projecting them to the same RGB image and then doing PCA compression. In the last 2 columns we show the estimated and ground truth trajectories. The top three rows show our results on CARLA; the
bottom three rows show our KITTI results.}
  \label{fig:tracking_quali}
\end{figure}

We show quantitative tracking results for our model and baselines on the  CARLA and KITTI datasets in Table \ref{tab:track_quant}. Figure \ref{fig:tracking_quali} visualizes  3D object trajectories over time from an overhead view. 
Note that we do not use any locality search, i.e., each model searches across the whole target frame for the configuration of the object.
Our model achieves superior performance on 3D object tracking than all the baselines. It dramatically  outperforms \basone{}, which highlights the importance of high spatial resolution for 3D feature learning and correspondence. It also outperforms \bastwo{}, which highlights the importance of amodal completion supported by our model. Lastly, all 3D methods outperform \basthree{} which highlights the effectiveness of learning features in a metric 3D feature space.

\begin{table}[h!]
\centering
\begin{tabular}{ |c|c|c| } 
\hline
Method & CARLA & KITTI\\
\hline
\model{} (ours) & \textbf{0.61} & \textbf{0.54}\\
\bastwo{} \cite{xie2020pointcontrast} & 0.55 & 0.48\\
%3D siamese + cosine(supervised) \cite{bertinetto2016fully} & 0.44 & 0.41\\
\basone{}  \cite{harley2020tracking} & 0.37 & 0.23\\ 
\basthree{} \cite{pmlr-v87-florence18a} & 0.24 & 0.19\\
\hline
\end{tabular}
\caption{\textbf{3D object tracking mean IOU across timesteps.}}
\label{tab:track_quant}
\end{table}

\begin{table}[h!]
\centering
\begin{tabular}{ |c|c|c|c|c| } 
\hline
Method & 0.25 & 0.30 & 0.40 & 0.50\\
\hline
\votenet{} \cite{qi2019deep} & 0.32 & 0.26 & 0.24 & 0.20\\ 
\ourvotenet{} (Ours) & \textbf{0.51} & \textbf{0.47} & \textbf{0.41} & \textbf{0.32}\\
\hline
\end{tabular}
\caption{\textbf{Comparing \votenet{} with \ourvotenet{} on mAP for different IOUs}. Pre-training the 3D point cloud features with \model{} significantly boosts performance.}
\label{table:votenet}
\end{table}

\subsection{Supervised 3D object detection in point clouds}
\label{sec:objdet_exp}

\paragraph{Setup}
We test the 3D feature representations obtained by \model{} in their ability to boost the performance of a supervised 3D object detector when used as initialization. We use  \votenet{} \cite{qi2019deep}, a state-of-the-art 3D object detector that combines the classic ideas of Hough voting in a deep neural network architecture that computes point features,  uses them to vote for locations and sizes of the 3D object boxes, and aggregates the votes to propose 3D object bounding boxes with associated confidence, in an end-to-end differentiable framework. 
%useful initial features for trainign a state-of-the-art 3D point cloud object detector of \cite{qi2019deep}
%warm-start supervised 3D object 
We modify \votenet{} \cite{qi2019deep} by replacing their point cloud feature extractor backbone with a trained  \singlemodel; we will refer to this as \ourvotenet{}. 
%We train \model{} backbone using 3D metric learning (Section \ref{sec:ml3d}) \todo{Explain it in model and refer it here}. Rest of the modules, experiment setup, and hyperparameters remain exactly same for both the models. 
The original backbone in \votenet{} uses PointNet++ \cite{qi2017pointnet++} to compute point features. %, \ourvotenet{} uses implicit functions, as described in Section \ref{sec:model}, to obtain point features. 
Both the \votenet{} backbone and \ourvotenet{} backbone sample 1024 points from the dense point cloud using farthest point sampling. The subsequent voting and proposal modules of \votenet{} then operate on these sampled point features to make the final predictions. In both cases, upon initialization, the 3D point features are trained end-to-end supervised for the 3D object detection task. We show below that our pre-training gives a significant boost to 3D object detection performance. 

\vspace{-0.1in}
\paragraph{Evaluation and comparison with baselines}
%Let \ourvotenet{} denote the Hough Voting network with our model as the 3D point cloud feature extractor, and let  \votenet{} denote the original model where point features are trained from scratch using PointNet++.

%We modify \votenet{} \cite{qi2019deep} by replacing their `point cloud feature extractor backbone` with \model, a model that we will refer to as \ourvotenet{}. We pre-train \model{} backbone using 3D metric learning (Section \ref{sec:ml3d}) \todo{Explain it in model and refer it here}. Rest of the modules, experiment setup, and hyperparameters remain exactly same for both the models. Where the original backbone in \votenet{} uses PointNet++ \cite{qi2017pointnet++} to get the point features, \ourvotenet{} uses implicit functions, as described in Section \ref{sec:model}, to get the point features. Both the \votenet{} and \ourvotenet{} backbone samples 1024 points and their features from the dense point cloud using farthest point sampling. The subsequent voting and proposal modules of \votenet{} then operate on these sampled point features to make the final predictions. 

We evaluate both  models on the CARLA dataset in the task of predicting 3D axis aligned car bounding boxes. 
%We assume all vehicles belonging to a single semantic class and therefore do not predict the class for detected objects. 
We show the mean average precision (mAP) scores for both the models for different IoU thresholds in Table \ref{table:votenet}.  \ourvotenet{} outperforms \votenet{} across all IoU thresholds.

%\subsection{3D object pose estimation}
%Given two viewpoints of the same object, we use RANSCA 

\subsection{Cross-view and cross-object 3D alignment}
\label{sec:corres_exp}

\paragraph{Setup}
We evaluate the correspondability of our 3D feature representations by testing how well they can align different views of the same instance, and different instances that belong to the same object category. Given the amodal 3D feature point clouds extracted for two objects in random viewpoints, we use orthogonal Procrustes problem with RANSAC  to estimate the 3D rigid body transformation that aligns them, similar to tracking a car in Section \ref{sec:tracking_exp}. 
Estimating fine grained correspondences between objects, once their rigid alignment has been estimated, is obtained using iterative closest point method \cite{Armesto10icra}, and we show such results in the supplementary file.

\paragraph{Evaluation and comparison with baselines}
We investigate two different setups:
i) Given two viewpoints of the same object, our goal is to estimate the relative 3D rotation between them.
ii) Given two objects of the same category in random viewpoints, our goal is to estimate a rigid 3d alignment between them. 
We compare our model against baselines \basone{} and \bastwo{}, for which we use RANSAC on voxel and point features, respectively, to estimate a 3D rigid transformation for the two objects.%, and mauch the similar way as done in Secftion    
%In this experiment we evaluate the correspondability of learnt features from our model and the baselines. Given single view RGB-D images of 2 object instances within a category and their corresponding 3D ground truth bounding box, we evaluate the accuracy of being able to retreive, their relative poses using RANSAC. We first infer the object-level features for each of the approaches, and then we use these inferred features, to calculate the cross correlation. 
%For \model{} and \cite{xie2020pointcontrast}, we infer the point features, whereas for \cite{adam3d} we infer voxel features. We then apply RANSAC on these cross correlations to find a rigid body transformation that satisfies most of these scores. 
\begin{table}[h!]
\centering
\begin{tabular}{ |c|c|c| } 
\hline
Method & cross-object & cross-view\\
\hline
\basone{}~\cite{harley2020tracking} & \textbf{0.18} & 0.21\\ 
\bastwo{}~\cite{xie2020pointcontrast} & 0.09 & 0.14\\
\model{} (Ours) & \textbf{0.18} & \textbf{0.58}\\
\hline
\end{tabular}
\caption{\textbf{Cross-object and cross-view 3D alignment accuracies in Shapenet dataset} (mean over 4 classes: Aeroplane, Mug, Car, Chair).}
\label{tab:correspondence}
\end{table}

The inferred 3D alignment is considered  correct if each of the yaw, roll, and pitch angles are within 10$^{\circ}$ of their respective ground truth values. In Table \ref{tab:correspondence} we show the cross-view (same-object) and cross-object 3D alignment accuracy for our  model and the baselines.
Our model performs the same as \basone{} on cross-object alignment, and outperforms both models on cross-view alignment. % the baselines, and the ordering of the models is similar to the 3D object tracking setup. 
Note that cross-object alignment is much harder than cross-view for all three models.

% We evaluate how correspondable the features learned by \model{} are. Given the features extracted for two entities, $\mathcal{E}_1$ and $\mathcal{E}_2$, in random poses, we would like the features at corresponding locations to be closer to each other as compared to features extracted from any other location. We compare our model on this task with Harley \etal's \cite{adam3d} voxel based features, which are optimized using a combination of RGB view prediction, occupancy prediction, and 3D contrastive losses. We, on the other hand, train our features using just 3D contrastive loss. We evaluate bothen   the models on the task of retrieving the pose betwe$\mathcal{E}_1$ and $\mathcal{E}_2$. Given two entities, we find the cross-correlation between their features. For Harley \etal's model, we crop the features corresponding to the entities from their learned 3D feature grid, and use them for calculating cross-correlation. For \model, we take the point cloud corresponding to the entities, calculate their point features using implicit functions, and use them for calculating cross-correlation.

\subsection{Generalization in view prediction}
We evaluate \model-\textit{OccRGB} in its ability to predict plausible images in scenes with novel number of objects, novel object appearance, and novel arrangements. We compare against  state-of-the-art view prediction models.  We further evaluate its ability for 3D occupancy prediction and completion. Extensive results of occupancy and RGB predictions on the ShapeNet dataset for our model can be found in Sectionn \ref{sec:nerf_exp}.
%These losses are applied only on  visible 3D points in target viewpoints clouds. 
% to be similar to other voxel implicit occupancy models of \cite{popov2020corenet,peng2020convolutional}.

We compare \model{} in RGB prediction against generative query network (GQN) \cite{Eslami1204} and geometry-aware recurrent networks (GRNN) \cite{commonsense}. We adapted GQN code to use posed RGB-D  images as input. GQN uses a 2D latent feature space to encode the RGB-D input image, and GRNN uses a discrete latent 3D feature map  of the scene. Our model uses a grid of functions and does not suffer from resolution limitations of the 3D feature map.

\begin{figure}[h!]
  \centering
  \includegraphics[width=0.48\textwidth]{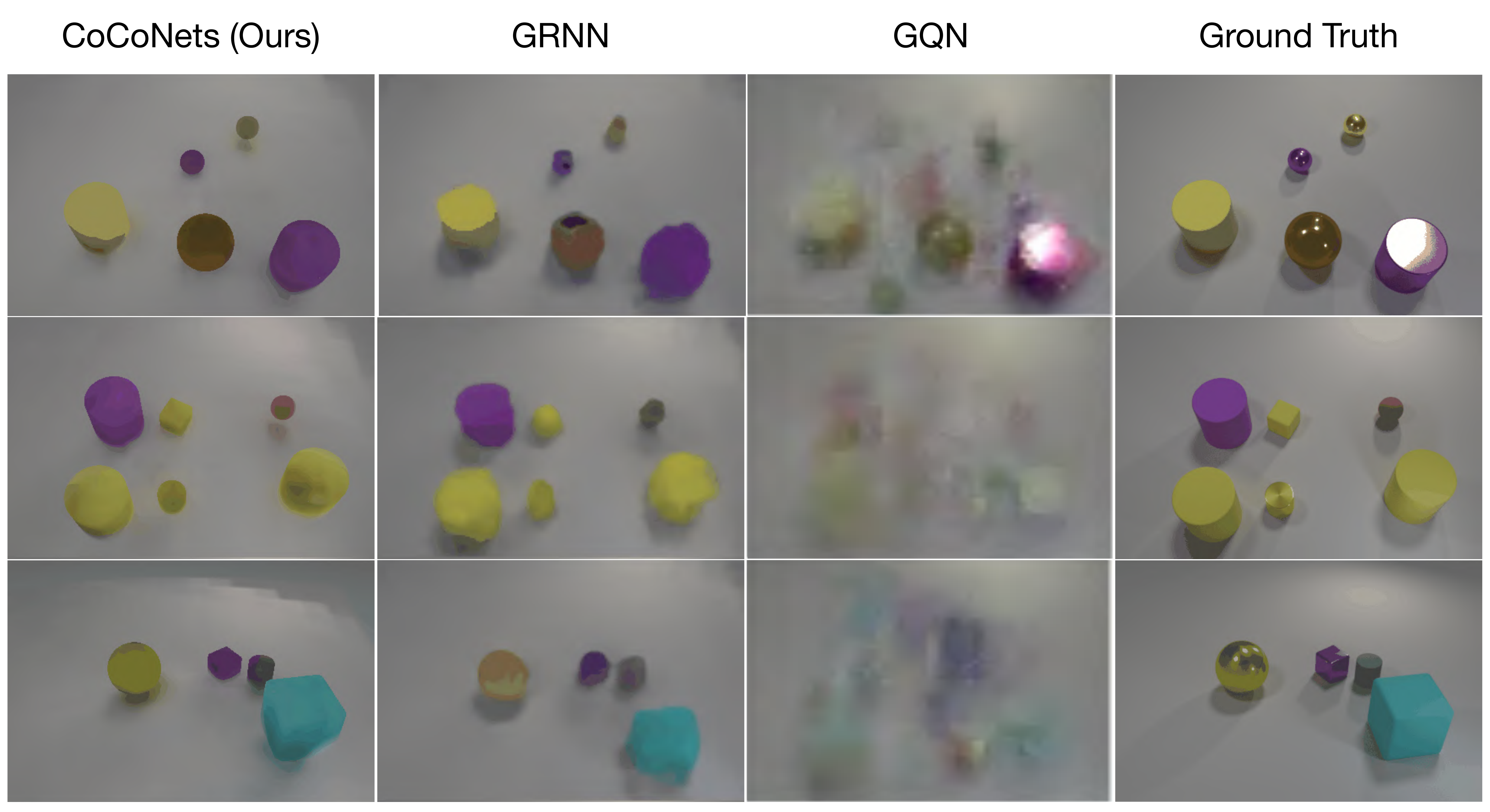}
  \caption{ 
    \textbf{Novel view prediction.} We compare  \model, GQN \cite{Eslami1204}, and GRNN \cite{commonsense} on the CLEVR \cite{johnson2017clevr} dataset. 
    %\model{} gives the most accurate renderings.
    }
  \label{fig:viewpred}
\end{figure}

%\model, on the other hand, also learns a 3D representation of the world, but instead of learning voxels, and thereby suffering from its drawbacks, it learns point features. 
\model{} predicts RGB images for target viewpoints as described in Sec.~\ref{sec:model1}. 
Figure \ref{fig:viewpred} shows the view generation results for \model, GQN, and GRNN. 
For qualitative results on RGB prediction and 3D occupancy prediction, details on the datasets used, architectural details and video results of our self-supervised tracking, please see our supplementary file.
%\model{} achieves better view predictions than the baselines.

\begin{figure}[h!]
  \centering
  \includegraphics[width=0.45\textwidth]{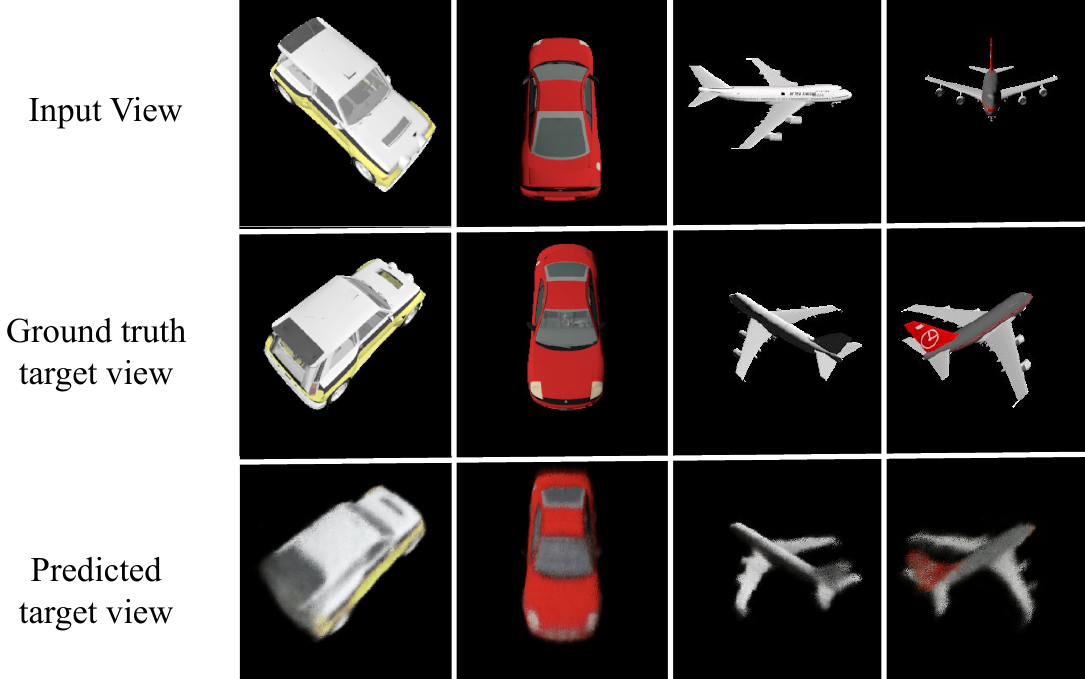}
  \caption{ 
    \textbf{RGB view regression} on the ShapeNet dataset using \model$^{NoDepthRGB}$. See the supplementary file for details on the experimental setup. 
    %\model{} gives the most accurate renderings.
    }
  \label{fig:rgbnerf}
\end{figure}

\paragraph{Limitations/Extensions}
\model{} can be extended to operate without the availability of depth maps at either train or test time using a differential rendering module \cite{sitzmann2020scene}, as shown in Figure~\ref{fig:rgbnerf}. %\model$^{NoDepthRGB}$. 
 Our supplementary material further contains results from variations of our model that do not assume depth available at test time, yet use it at training time.

% \vspace{-0.1in}
\section{Conclusion}
% \vspace{-0.1in}
% \thispagestyle{empty}
We present a method for learning 3D  visual feature representations by self-supervised view and depth prediction from posed RGB and RGB-D images. Our networks lift input 2.5D images of objects and scenes into latent thee-dimensional function grids, which can be decoded to infinite-resolution 3D  occupancy and 3D feature predictions of the object or scene. Our networks are trained by predicting views using a contrastive mutual information maximization objective. We evaluate the emergent 3D  visual feature  representations in 3D object tracking in dynamic scenes and in cross-view and cross-object alignment estimation. We empirically demonstrate that the features are semantically meaningful and outperform popular point-based supervision \cite{xie2020pointcontrast} which does not consider 3D completion, and discrete voxel 3D latent feature maps of previous works \cite{adam3d,harley2020tracking} which are limited by the spatial resolution of the 3D feature grid. Moreover, our models can better generalize to novel scenes with unseen number and appearance of objects than networks that do not encode 3D structure \cite{Eslami1204} or do not incorporate 3D convolutional  modules \cite{sitzmann2020scene}. They make a step towards self-supervised learning of amodal 3D feature representations, which we show are useful for 3D object tracking and correspondence. %, but potentially to other tasks that care about free space and occupancy inference, such as object manipulation. 
Avenues for future work include learning such representations directly from dynamic videos, and relaxing the egomotion and depth supervision.%  are avenues for our future work. % in 3D generative models. 

%We showed their usefule

%egostabilization operations generate compelling RGb images under conbinatorial generalization of scene appearance  % can be inferred alognside  feature vectors for any sampled point clouds can be sampled  

%Convolutional Continuous Constrastive Networks for view and depth prediction  \model, a model 
\section{Acknowledgement}
This work has been funded by  Sony AI,  DARPA Machine Common Sense, a NSF CAREER award, the Air Force Office of Scientific Research under award number FA9550-20-1-0423. Any opinions, findings, and conclusions or recommendations expressed in this material are those of the author(s) and do not necessarily reflect the views of the United States Air Force.

\clearpage
{\small
\bibliographystyle{ieee_fullname}
\bibliography{refsnew,refs3dqnets}

\begin{thebibliography}{10}\itemsep=-1pt

\bibitem{Armesto10icra}
Leopoldo Armesto, Javier Minguez, and Luis Montesano.
\newblock A generalization of the metric-based iterative closest point
  technique for 3d scan matching.
\newblock In {\em 2010 IEEE International Conference on Robotics and
  Automation}, pages 1367--1372. IEEE, 2010.

\bibitem{brock2019large}
Andrew Brock, Jeff Donahue, and Karen Simonyan.
\newblock Large scale gan training for high fidelity natural image synthesis,
  2019.

\bibitem{shapenet2015}
Angel~X. Chang, Thomas Funkhouser, Leonidas Guibas, Pat Hanrahan, Qixing Huang,
  Zimo Li, Silvio Savarese, Manolis Savva, Shuran Song, Hao Su, Jianxiong Xiao,
  Li Yi, and Fisher Yu.
\newblock {ShapeNet: An Information-Rich 3D Model Repository}.
\newblock {\em arXiv:1512.03012}, 2015.

\bibitem{chen2019learning}
Zhiqin Chen and Hao Zhang.
\newblock Learning implicit fields for generative shape modeling.
\newblock In {\em Proceedings of the IEEE/CVF Conference on Computer Vision and
  Pattern Recognition}, pages 5939--5948, 2019.

\bibitem{chibane2020implicit}
Julian Chibane, Thiemo Alldieck, and Gerard Pons-Moll.
\newblock Implicit functions in feature space for 3d shape reconstruction and
  completion, 2020.

\bibitem{Dosovitskiy17}
Alexey Dosovitskiy, German Ros, Felipe Codevilla, Antonio Lopez, and Vladlen
  Koltun.
\newblock {CARLA}: {An} open urban driving simulator.
\newblock In {\em CORL}, pages 1--16, 2017.

\bibitem{Eslami1204}
S.~M.~Ali Eslami, Danilo Jimenez~Rezende, Frederic Besse, Fabio Viola, Ari~S.
  Morcos, Marta Garnelo, Avraham Ruderman, Andrei~A. Rusu, Ivo Danihelka, Karol
  Gregor, David~P. Reichert, Lars Buesing, Theophane Weber, Oriol Vinyals, Dan
  Rosenbaum, Neil Rabinowitz, Helen King, Chloe Hillier, Matt Botvinick, Daan
  Wierstra, Koray Kavukcuoglu, and Demis Hassabis.
\newblock Neural scene representation and rendering.
\newblock {\em Science}, 360(6394):1204--1210, 2018.

\bibitem{pmlr-v87-florence18a}
Peter~R. Florence, Lucas Manuelli, and Russ Tedrake.
\newblock Dense object nets: Learning dense visual object descriptors by and
  for robotic manipulation.
\newblock In Aude Billard, Anca Dragan, Jan Peters, and Jun Morimoto, editors,
  {\em Proceedings of The 2nd Conference on Robot Learning}, volume~87 of {\em
  Proceedings of Machine Learning Research}, pages 373--385. PMLR, 29--31 Oct
  2018.

\bibitem{Geiger2013IJRR}
Andreas Geiger, Philip Lenz, Christoph Stiller, and Raquel Urtasun.
\newblock Vision meets robotics: The kitti dataset.
\newblock {\em International Journal of Robotics Research (IJRR)}, 2013.

\bibitem{Geiger2012CVPR}
Andreas Geiger, Philip Lenz, and Raquel Urtasun.
\newblock Are we ready for autonomous driving? {T}he {KITTI} vision benchmark
  suite.
\newblock In {\em CVPR}, 2012.

\bibitem{genova2020local}
Kyle Genova, Forrester Cole, Avneesh Sud, Aaron Sarna, and Thomas Funkhouser.
\newblock Local deep implicit functions for 3d shape, 2020.

\bibitem{DBLP:journals/corr/abs-1906-02739}
Georgia Gkioxari, Jitendra Malik, and Justin Johnson.
\newblock Mesh {R-CNN}.
\newblock {\em CoRR}, abs/1906.02739, 2019.

\bibitem{NIPS2014_5423}
Ian Goodfellow, Jean Pouget-Abadie, Mehdi Mirza, Bing Xu, David Warde-Farley,
  Sherjil Ozair, Aaron Courville, and Yoshua Bengio.
\newblock Generative adversarial nets.
\newblock In Z. Ghahramani, M. Welling, C. Cortes, N.~D. Lawrence, and K.~Q.
  Weinberger, editors, {\em NIPS}, pages 2672--2680, 2014.

\bibitem{han2019video}
Tengda Han, Weidi Xie, and Andrew Zisserman.
\newblock Video representation learning by dense predictive coding, 2019.

\bibitem{harley2020tracking}
Adam~W Harley, Shrinidhi~K Lakshmikanth, Paul Schydlo, and Katerina
  Fragkiadaki.
\newblock Tracking emerges by looking around static scenes, with neural 3d
  mapping.
\newblock {\em arXiv preprint arXiv:2008.01295}, 2020.

\bibitem{adam3d}
Adam~W Harley, Fangyu Li, Shrinidhi~K Lakshmikanth, Xian Zhou, Hsiao-Yu~Fish
  Tung, and Katerina Fragkiadaki.
\newblock Learning from unlabelled videos using contrastive predictive neural
  3d mapping.
\newblock In {\em ICLR}, 2020.

\bibitem{hartley2003multiple}
Richard Hartley and Andrew Zisserman.
\newblock {\em Multiple view geometry in computer vision}.
\newblock Cambridge university press, 2003.

\bibitem{he2020momentum}
Kaiming He, Haoqi Fan, Yuxin Wu, Saining Xie, and Ross Girshick.
\newblock Momentum contrast for unsupervised visual representation learning,
  2020.

\bibitem{he2016deep}
Kaiming He, Xiangyu Zhang, Shaoqing Ren, and Jian Sun.
\newblock Deep residual learning for image recognition.
\newblock In {\em Proceedings of the IEEE conference on computer vision and
  pattern recognition}, pages 770--778, 2016.

\bibitem{johnson2017clevr}
Justin Johnson, Bharath Hariharan, Laurens van~der Maaten, Li Fei-Fei, C
  Lawrence~Zitnick, and Ross Girshick.
\newblock {CLEVR}: A diagnostic dataset for compositional language and
  elementary visual reasoning.
\newblock In {\em Proceedings of the IEEE Conference on Computer Vision and
  Pattern Recognition}, pages 2901--2910, 2017.

\bibitem{LSM}
Abhishek Kar, Christian H{\"{a}}ne, and Jitendra Malik.
\newblock Learning a multi-view stereo machine.
\newblock In {\em NIPS}, 2017.

\bibitem{DBLP:journals/corr/abs-1711-07566}
Hiroharu Kato, Yoshitaka Ushiku, and Tatsuya Harada.
\newblock Neural 3d mesh renderer.
\newblock {\em CoRR}, abs/1711.07566, 2017.

\bibitem{kingma2018glow}
Diederik~P. Kingma and Prafulla Dhariwal.
\newblock Glow: Generative flow with invertible 1x1 convolutions, 2018.

\bibitem{kingma2013auto}
Diederik~P Kingma and Max Welling.
\newblock Auto-encoding variational bayes.
\newblock {\em arXiv:1312.6114}, 2013.

\bibitem{le2020novel}
Hoang-An Le, Thomas Mensink, Partha Das, and Theo Gevers.
\newblock Novel view synthesis from single images via point cloud
  transformation, 2020.

\bibitem{lin2017learning}
Chen-Hsuan Lin, Chen Kong, and Simon Lucey.
\newblock Learning efficient point cloud generation for dense 3d object
  reconstruction, 2017.

\bibitem{10.1145/37402.37422}
William~E. Lorensen and Harvey~E. Cline.
\newblock Marching cubes: A high resolution 3d surface construction algorithm.
\newblock {\em SIGGRAPH Comput. Graph.}, 21(4):163–169, Aug. 1987.

\bibitem{mescheder2019occupancy}
Lars Mescheder, Michael Oechsle, Michael Niemeyer, Sebastian Nowozin, and
  Andreas Geiger.
\newblock Occupancy networks: Learning 3d reconstruction in function space.
\newblock In {\em Proceedings of the IEEE/CVF Conference on Computer Vision and
  Pattern Recognition}, pages 4460--4470, 2019.

\bibitem{mildenhall2020nerf}
Ben Mildenhall, Pratul~P. Srinivasan, Matthew Tancik, Jonathan~T. Barron, Ravi
  Ramamoorthi, and Ren Ng.
\newblock Nerf: Representing scenes as neural radiance fields for view
  synthesis, 2020.

\bibitem{doi:10.1177/2041669518788887}
Bence Nanay.
\newblock The importance of amodal completion in everyday perception.
\newblock {\em i-Perception}, 9(4):2041669518788887, 2018.
\newblock PMID: 30109014.

\bibitem{nguyen2019hologan}
Thu Nguyen-Phuoc, Chuan Li, Lucas Theis, Christian Richardt, and Yong-Liang
  Yang.
\newblock Hologan: Unsupervised learning of 3d representations from natural
  images.
\newblock In {\em Proceedings of the IEEE/CVF International Conference on
  Computer Vision}, pages 7588--7597, 2019.

\bibitem{DBLP:journals/corr/NovotnyLV17a}
David Novotn{\'{y}}, Diane Larlus, and Andrea Vedaldi.
\newblock Learning 3d object categories by looking around them.
\newblock {\em CoRR}, abs/1705.03951, 2017.

\bibitem{novotny2020canonical}
David Novotny, Roman Shapovalov, and Andrea Vedaldi.
\newblock Canonical 3d deformer maps: Unifying parametric and non-parametric
  methods for dense weakly-supervised category reconstruction, 2020.

\bibitem{park2019deepsdf}
Jeong~Joon Park, Peter Florence, Julian Straub, Richard Newcombe, and Steven
  Lovegrove.
\newblock Deepsdf: Learning continuous signed distance functions for shape
  representation.
\newblock In {\em Proceedings of the IEEE/CVF Conference on Computer Vision and
  Pattern Recognition}, pages 165--174, 2019.

\bibitem{peng2020convolutional}
Songyou Peng, Michael Niemeyer, Lars Mescheder, Marc Pollefeys, and Andreas
  Geiger.
\newblock Convolutional occupancy networks, 2020.

\bibitem{popov2020corenet}
Stefan Popov, Pablo Bauszat, and Vittorio Ferrari.
\newblock Corenet: Coherent 3d scene reconstruction from a single rgb image,
  2020.

\bibitem{qi2019deep}
Charles~R Qi, Or Litany, Kaiming He, and Leonidas~J Guibas.
\newblock Deep hough voting for 3d object detection in point clouds.
\newblock In {\em Proceedings of the IEEE International Conference on Computer
  Vision}, pages 9277--9286, 2019.

\bibitem{qi2017pointnet++}
Charles~Ruizhongtai Qi, Li Yi, Hao Su, and Leonidas~J Guibas.
\newblock Pointnet++: Deep hierarchical feature learning on point sets in a
  metric space.
\newblock In {\em Advances in neural information processing systems}, pages
  5099--5108, 2017.

\bibitem{riegler2020free}
Gernot Riegler and Vladlen Koltun.
\newblock Free view synthesis, 2020.

\bibitem{DBLP:journals/corr/abs-1905-05172}
Shunsuke Saito, Zeng Huang, Ryota Natsume, Shigeo Morishima, Angjoo Kanazawa,
  and Hao Li.
\newblock {PIFu}: {P}ixel-aligned implicit function for high-resolution clothed
  human digitization.
\newblock {\em arXiv:1905.05172}, 2019.

\bibitem{schonemann1966generalized}
Peter~H Sch{\"o}nemann.
\newblock A generalized solution of the orthogonal procrustes problem.
\newblock {\em Psychometrika}, 31(1):1--10, 1966.

\bibitem{schwarz2020graf}
Katja Schwarz, Yiyi Liao, Michael Niemeyer, and Andreas Geiger.
\newblock Graf: Generative radiance fields for 3d-aware image synthesis.
\newblock {\em arXiv preprint arXiv:2007.02442}, 2020.

\bibitem{sitzmann2018deepvoxels}
Vincent Sitzmann, Justus Thies, Felix Heide, Matthias Nie{\ss}ner, Gordon
  Wetzstein, and Michael Zollh{\"o}fer.
\newblock Deep{V}oxels: {L}earning persistent {3D} feature embeddings.
\newblock In {\em CVPR}, 2019.

\bibitem{sitzmann2020scene}
Vincent Sitzmann, Michael Zollhöfer, and Gordon Wetzstein.
\newblock Scene representation networks: Continuous 3d-structure-aware neural
  scene representations, 2020.

\bibitem{DBLP:journals/corr/TatarchenkoDB15}
Maxim Tatarchenko, Alexey Dosovitskiy, and Thomas Brox.
\newblock Single-view to multi-view: {R}econstructing unseen views with a
  convolutional network.
\newblock In {\em ECCV}, 2016.

\bibitem{tewari2020state}
Ayush Tewari, Ohad Fried, Justus Thies, Vincent Sitzmann, Stephen Lombardi,
  Kalyan Sunkavalli, Ricardo Martin-Brualla, Tomas Simon, Jason Saragih,
  Matthias Nießner, Rohit Pandey, Sean Fanello, Gordon Wetzstein, Jun-Yan Zhu,
  Christian Theobalt, Maneesh Agrawala, Eli Shechtman, Dan~B Goldman, and
  Michael Zollhöfer.
\newblock State of the art on neural rendering, 2020.

\bibitem{tobin2019geometryaware}
Josh Tobin, OpenAI Robotics, and Pieter Abbeel.
\newblock Geometry-aware neural rendering, 2019.

\bibitem{TulGupFouEfrMal17}
Shubham Tulsiani, Saurabh Gupta, David~F. Fouhey, Alexei~A. Efros, and Jitendra
  Malik.
\newblock Factoring shape, pose, and layout from the 2d image of a 3d scene.
\newblock {\em CoRR}, abs/1712.01812, 2017.

\bibitem{DBLP:journals/corr/TulsianiZEM17}
Shubham Tulsiani, Tinghui Zhou, Alexei~A. Efros, and Jitendra Malik.
\newblock Multi-view supervision for single-view reconstruction via
  differentiable ray consistency.
\newblock {\em CoRR}, abs/1704.06254, 2017.

\bibitem{commonsense}
Hsiao-Yu~Fish Tung, Ricson Cheng, and Katerina Fragkiadaki.
\newblock Learning spatial common sense with geometry-aware recurrent networks.
\newblock In {\em CVPR}, 2019.

\bibitem{van2016conditional}
Aaron van~den Oord, Nal Kalchbrenner, Lasse Espeholt, Oriol Vinyals, Alex
  Graves, et~al.
\newblock Conditional image generation with pixelcnn decoders.
\newblock In {\em NIPS}, pages 4790--4798, 2016.

\bibitem{oord2018representation}
Aaron van~den Oord, Yazhe Li, and Oriol Vinyals.
\newblock Representation learning with contrastive predictive coding.
\newblock {\em arXiv:1807.03748}, 2018.

\bibitem{Wiles_2020_CVPR}
Olivia Wiles, Georgia Gkioxari, Richard Szeliski, and Justin Johnson.
\newblock Synsin: End-to-end view synthesis from a single image.
\newblock In {\em Proceedings of the IEEE/CVF Conference on Computer Vision and
  Pattern Recognition (CVPR)}, June 2020.

\bibitem{xie2020pointcontrast}
Saining Xie, Jiatao Gu, Demi Guo, Charles~R. Qi, Leonidas~J. Guibas, and Or
  Litany.
\newblock Pointcontrast: Unsupervised pre-training for 3d point cloud
  understanding, 2020.

\bibitem{yan2017perspective}
Xinchen Yan, Jimei Yang, Ersin Yumer, Yijie Guo, and Honglak Lee.
\newblock Perspective transformer nets: Learning single-view 3d object
  reconstruction without 3d supervision, 2017.

\bibitem{you2020keypointnet}
Yang You, Yujing Lou, Chengkun Li, Zhoujun Cheng, Liangwei Li, Lizhuang Ma,
  Cewu Lu, and Weiming Wang.
\newblock Keypointnet: A large-scale 3d keypoint dataset aggregated from
  numerous human annotations.
\newblock {\em arXiv preprint arXiv:2002.12687}, 2020.

\bibitem{yu2020pixelnerf}
Alex Yu, Vickie Ye, Matthew Tancik, and Angjoo Kanazawa.
\newblock pixelnerf: Neural radiance fields from one or few images, 2020.

\end{thebibliography}
}
\clearpage
\section*{Supplementary File}

% \section{Overview}
%\todo{Write overview here}
% In this supplementary file, we provide additional details on the datasets used to train and test the models (Sec.~\ref{sec:dataprep}), implementation details (Sec.~\ref{sec:impl}), and we provide a more-detailed analysis of the results (Sec.~\ref{sec:results}).
The structure of this supplementary file is as follows: Section~\ref{sec:dataprep} covers the details on dataset preparation. Section~\ref{sec:impl} elucidates the implementation details, including inputs to our model, architecture details, and hyperparameter values used. Finally, Section~\ref{sec:results} provides a more detailed analysis of the results. 
% \sout{We also include a video named `suppl\_10375.m4v` with additional results and urge the reviewers to refer that as well.}

\section{Dataset Preparation}\label{sec:dataprep}
\paragraph{\normalfont \textbf{CARLA dataset.}} 
%\todo{Write about both tracking and multiview dataset here}
We use the CARLA simulator \cite{Dosovitskiy17} to generate multi-view data for training, and tracking data for testing. CARLA is an open-source simulator for urban driving scenes, which allows  multiple cameras to be placed at arbitrary positions. We generate  episodes 300 frames long, with a framerate of 10 FPS. Each episode is captured from 6 random camera locations, sampled from a set of 18 cameras that lie on a hemisphere. We treat Towns 0-3 as the training set, and Towns 4-5 as the test set. In total, we use 388 scenes for self-supervised multi-view training and 188 scenes for tracking evaluation. The resolution of our RGB and depth images is 128x384.

\paragraph{\normalfont \textbf{KITTI dataset.}} 
We use the KITTI benchmark \cite{Geiger2012CVPR} to evaluate our model on the tracking task. We use the ``tracking" subset of KITTI to test our model. This subset has 8008 frames across twenty labelled sequences. 
We create 8-frame sub-sequences of this data. We make sure that all the eight frames have a valid label for the target
object. The egomotion information in the “tracking” data is only approximate.

\paragraph{\normalfont \textbf{Shapenet dataset.}}
We use the meshes from ShapeNetCore \cite{shapenet2015}. We first normalize the ShapeNet meshes. We then render RGB-D data for them using Blender, where we place 24 cameras around the object. The cameras are placed at a distance of 2m, with azimuth ranging from $0^{\circ}$ - $360^{\circ}$ at intervals of $45^{\circ}$, and elevation ranging from $0^{\circ}$ - $80^{\circ}$ at intervals of $20^{\circ}$.

\paragraph{\normalfont \textbf{CLEVR dataset.}} 
We build upon the CLEVR Blender simulator \cite{johnson2017clevr}. Our scenes consist of cubes, spheres, and cylinders. We create scenes as follows: each object model is randomly rotated ($0^{\circ}$ to $360^{\circ}$ along vertical axis), translated (randomly within a sphere of radius 10.5 units) and scaled (0.25 to 1.25 times the actual size). Each scene contains $2-10$ objects. We  randomly vary the lighting of each scene. We render each scene by placing 28 RGB-D cameras at elevations ranging from $26^{\circ}$  to $80^{\circ}$ with $13^{\circ}$ increments and azimuths ranging from $0^{\circ}$ to $360^{\circ}$  with $45^{\circ}$ increments. Each camera is placed within a sphere of radius 1.5 metres from the center of the scene.

\section{Implementation details}\label{sec:impl}
\paragraph{\normalfont \textbf{Code, training details and computation complexity.}}
Our model is implemented in Python/Pytorch. We use a batch size of 2 and our learning rate is set at $10^{-4}$. We use the Adam optimizer with $\beta_1 = 0.9$, $\beta_2 = 0.999$.
Our model takes 24 hours (approximately 100k iterations) of training for convergence on a single V100 GPU. As far as the running time of the proposed method for tasks in the experiments is concerned, extraction of point features from an RGB-D image takes 0.3 seconds for our model on a V100 GPU. A single iteration of Object tracking (Section 4.1) takes 5 seconds, Object Detection (Section 4.2) takes 0.4 seconds, Object Alignment (Section 4.3) takes 6 seconds and View Prediction using \model-\textit{OccRGB} (Section 4.4) takes 0.6 seconds.

\paragraph{\normalfont \textbf{Inputs.}}
 We randomly select images from 2 views which includes the RGB-D and relative ground-truth egomotion  (query view and target view)  as input for training our model, while we use a single view for testing.

\paragraph{\normalfont \textbf{2.5D-to-3D lifting.}}
Our 2.5D-to-3D unprojection module takes as input RGB-D images and converts it into a 4D tensor $\textbf{U} \in \mathbb{R}^{w \times h \times d \times 4}$, where $w,h,d$ is $64,64,64$. We use perspective (un)projection to fill the 3D grid with samples from 2D image. Specifically, using pinhole camera model \cite{hartley2003multiple}, we find the floating-point 2D pixel location that every cell in the 3D grid, indexed by the coordinate $(i,j,k)$, projects onto from the current camera viewpoint. This is given by $[u,v]^T = \K \S [i,j,k]^T$, where $\S$, the similarity transform, converts memory coordinates to camera coordinates and $\K$, the camera intrinsics, convert camera coordinates to pixel coordinates.
Bilinear interpolation is applied on pixel values to fill the grid cells. We obtain a binary occupancy grid $\textbf{O} \in \mathbb{R}^{w \times h \times d \times 1}$ from the depth image $\textbf{D}$ in a similar way. This occupancy is then concatenated with the unprojected RGB to get a tensor $[\textbf{U}, \textbf{O}] \in \mathbb{R}^{w \times h \times d \times 4}$. This tensor is then passed through a 3D encoder-decoder network, the architecture of which is as follows: 4-2-64, 4-2-128, 4-2-256, 4-0.5-128, 4-0.5-64, 1-1-$F$. Here, we use the notation $k$-$s$-$c$ for kernel-stride-channels, and $F$ is the feature dimension, which we set to $F=32$.
We concatenate the output of transposed convolutions in decoder with same resolution feature map output from the encoder. The concatenated tensor is then passed to the next layer in the decoder. We use leaky ReLU activation and batch normalization after every convolution layer, except for the last one in each network. We obtain our 3D feature map $\textbf{M}$ as the output of this process. Even though we do not use multiple views as input in the paper, our system can handle that easily by orienting the feature map $\textbf{M}_v$ obtained from each view $v$ to a reference view to cancel the egomotion, and then fusing them to obtain the final feature map $\textbf{M}$.

\paragraph{\normalfont \textbf{Implicit function parameterization.}}
To predict the features, RGB, and occupancy values for different 3D physical points within a voxel grid, we use implicit functions parameterized by neural networks. As mentioned in the main paper, we denote the operation of obtaining point features, occupancies and colors by querying the feature map $\map$ at continuous locations $(X, Y, Z)$ by $\F^f\left(\map,(X,Y,Z)\right),\F^o\left(\map,(X,Y,Z)\right), \F^c\left(\map,(X,Y,Z)\right)$ respectively. We parameterize these three functions using neural networks with identical architectures, with the exception that the output of $\F^f$, $\F^o$, $\F^c$ is 32-D, 1-D and 3D, respectively. There is no weight sharing between these three neural networks. For brevity, we will use $\F$ when explaining properties common to all the three networks/functions.  

The architecture of $\F$ is similar to that of \cite{peng2020convolutional}. Given the point $(X, Y, Z)$ that we want to featurize, we first infer the trilinearly-interpolated feature vector at point $(X, Y, Z)$ from the 8 neighbouring voxel features, which we denote as $c$. We encode the coordinate $(X, Y, Z)$ into a 32-D feature vector using a linear layer. We denote this vector as $p$.
The inputs $c$ and $p$ are then processed as follows: 

\begin{multline}
out = RN_i( RN_{i-1}( \cdots RN_1( p + FC_1(c)) \\ \cdots] + FC_{i-1}(c)) + FC_{i}(c)).
\end{multline}
We set $i=5$ for our usecase. $FC_i$ is a linear layer which gives a 32-dimensional output. $RN_i$ is a 2 layer ResNet block \cite{he2016deep}. The architecture of ResNet block is as follows: ReLU, 32-32, ReLU, 32-32. Here, we use the notation $i_u-o_u$ to represent a linear layer, where $i_u$ is the input dimension, and $o_u$ is the output dimension of that layer. Each ResNet block finally generates the output by adding the input to the output of the above layers. $out$ is then passed through a ReLU activation function followed by a linear layer, the output dimension of which depends on the property of the point we are trying to predict.

\paragraph{\normalfont \textbf{\model{} on RGB images during inference for 3D alignment.}}
Our model can operate without depth maps at inference time. We achieve this by replacing our bottom-up 3D convolutional encoder-decoder architecture with a ResNet-18 \cite{he2016deep} which directly operates on RGB image from the target view $\Image_{target}$. Our ResNet-18 encodes $\Image_{target}$ to a 1D feature vector $\mathbf{x}_{target}$. During training, our top-down mapping of $(\Image_{inp},\D_{inp})$ to $\map_{inp}^{\mV}$ and then to feature cloud $\{   \left( X,Y,Z,\F(\map_{inp}^{\mV}, (X,Y,Z)) \right) \}$ remains the same,  we instead replace the bottom-up feature point clouds obtained from $\map_{target}$ with feature point cloud $\{   \left( X,Y,Z,\F(\mathbf{x}_{target}, (X,Y,Z)) \right) \}$ obtained by interpolating the 1D feature vector $\mathbf{x}_{target}$. We then train these feature clouds using the same contrastive loss discussed in Section \ref{sec:model}. Our bottom-up network gives us an option to operate without depth images during inference time. In Section \ref{sec:poseestimation}, we show the comparison on 3D alignment of our model operating on RGB versus RGB-D images.

\paragraph{\normalfont \textbf{Volumetric rendering using \model{} on RGB images.}}
\label{sec:nerf_exp}
We leverage our model's continuous representations on the task of rendering novel views on the ShapeNet dataset without depth as input and call it $\model^{NoDepthRGB}$. In this experiment, since we do not use depth information, we only pass the RGB image from the input view, $I_{inp}$, as an unprojected 3D tensor $U \in R^{w \times h \times d \times 3}$, to the top-down 3D-CNN encoder-decoder (Figure \ref{fig:model}(a)) to get $M_{inp}$. Then we take the camera pose change between the target (the view for which we want to render the image) and the input view, defined as $V$, and sample $w \times h$ number of rays, sampling 32 points on each ray, as per the method given in NeRF \cite{mildenhall2020nerf}. We use these sampled points, defined as $D_{target}^{V}$, to query $M_{inp}$ using trilinear interpolation. Thereafter, we use the MLP architecture given in \cite{mildenhall2020nerf} and pass as input the positional encoding of the query point, its viewing direction and the corresponding interpolated feature vector, getting $(r,g,b,\sigma)$ as the output for each query point. Extracted $(r,g,b,\sigma)$ of all query points on each ray are then passed into the volume rendering module of \cite{mildenhall2020nerf} to get the final RGB image, $\hat{I}_{target}$. We use MSE as the rendering loss between the ground truth RGB, $I_{target}$ and the RGB image rendered by NeRF, $\hat{I}_{target}$. Figure \ref{fig:rgbnerf} shows the qualitative results for this experiment. A concurrent work to this experiment is pixelNeRF \cite{yu2020pixelnerf}, with the difference being that we operate on a 3D feature tensor, whereas \cite{yu2020pixelnerf} operates on a 2D feature tensor.

\section{Results}\label{sec:results}
In this section, we show more results on occupancy prediction, RGB view prediction, dense correspondence, vehicle tracking and pose alignment when operating on RGB input.

\subsection{Occupancy prediction}
\begin{figure*}[h!]
  \centering
  \includegraphics[width=0.5\textwidth]{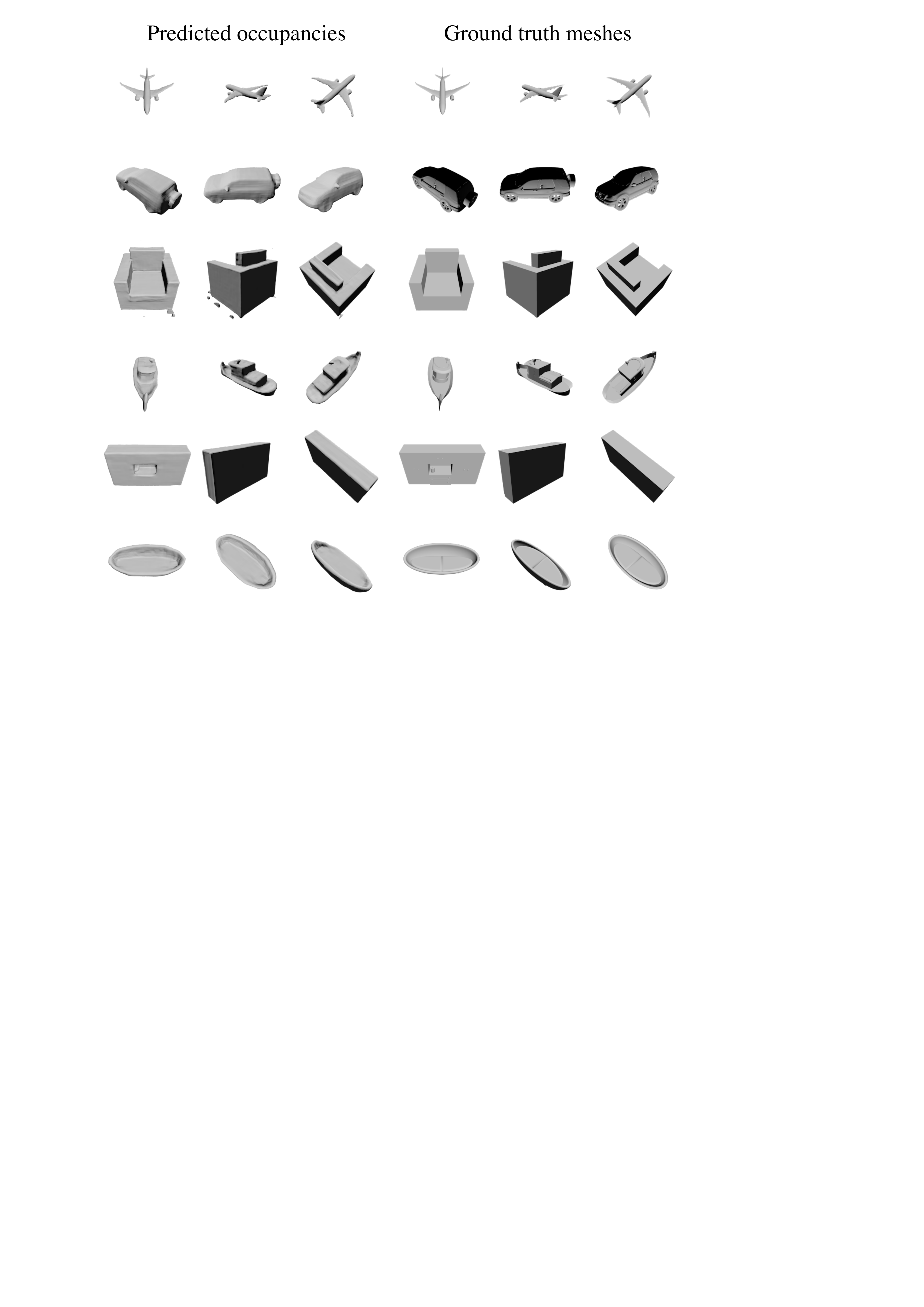}
  \caption{\textbf{Occupancies predicted by our model for ShapeNet objects.} First three columns show the renderings of the mesh created from predicted occupancies from three views. Last three columns show the same for ground truth mesh.}
  \label{fig:occpred1}
\end{figure*}

Figure~\ref{fig:occpred1} shows the occupancies predicted by our model on ShapeNet objects. The first three columns show the renderings of the mesh created from predicted occupancies from three views. The last three columns show the corresponding ground truth. We also show multiview renderings of the predicted occupancies and the ground truth meshes as GIFs in the supplementary video on our project page and urge the readers to refer that. At inference time, we extract meshes by applying Multiresolution IsoSurface
Extraction (MISE) \cite{mescheder2019occupancy}. Table \ref{table:occpred_iou} shows the IOU for occupancy prediction for our model on CARLA and ShapeNet datasets. For ShapeNet, we report the mean IOU over all the classes.

\begin{table}[h!]
\centering
\begin{tabular}{ |c|c| } 
\hline
Dataset & IOU \\
\hline
CARLA & 0.79\\ 
ShapeNet & 0.56\\
\hline
\end{tabular}
\caption{\textbf{Occupancy prediction evaluation.} Metric used is IOU.}
\label{table:occpred_iou}
\end{table}

\subsection{RGB view prediction}
\begin{figure*}[h!]
  \centering
  \includegraphics[width=0.5\textwidth]{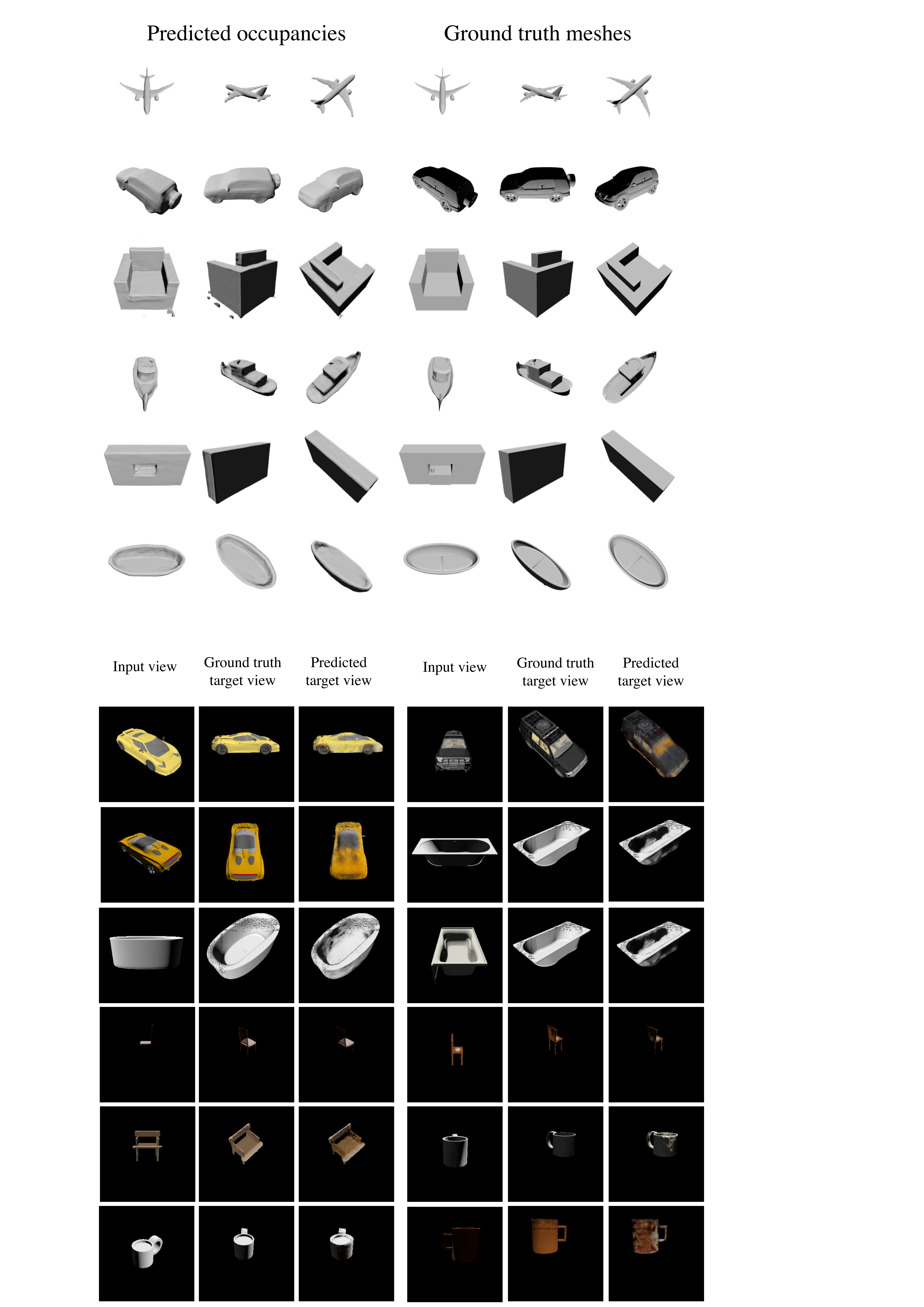}
  \caption{\textbf{Neural renders on ShapeNet dataset from \model-\textit{OccRGB}.} The first column shows the view given as input to our model. The second column shows the ground truth target view. The third column shows the predicted target view. Columns 4-6 follow the same convention.}
  \label{fig:rgbpred}
\end{figure*}

Figure~\ref{fig:rgbpred} shows qualitative results on the RGB novel view prediction task by our model on the ShapeNet dataset using depth as input. On the other hand, Figure \ref{fig:rgbnerf} shows the views rendered without using depth via the NeRF volumetric renderer.  Given an input view (first column), our model can render the scene from an arbitrary viewpoint (third column). We compare this rendering with the ground truth RGB for that view (second column).

\subsection{Self-supervised 3D object tracking}
\begin{figure*}[h!]
  \centering
  \includegraphics[width=0.72\textwidth]{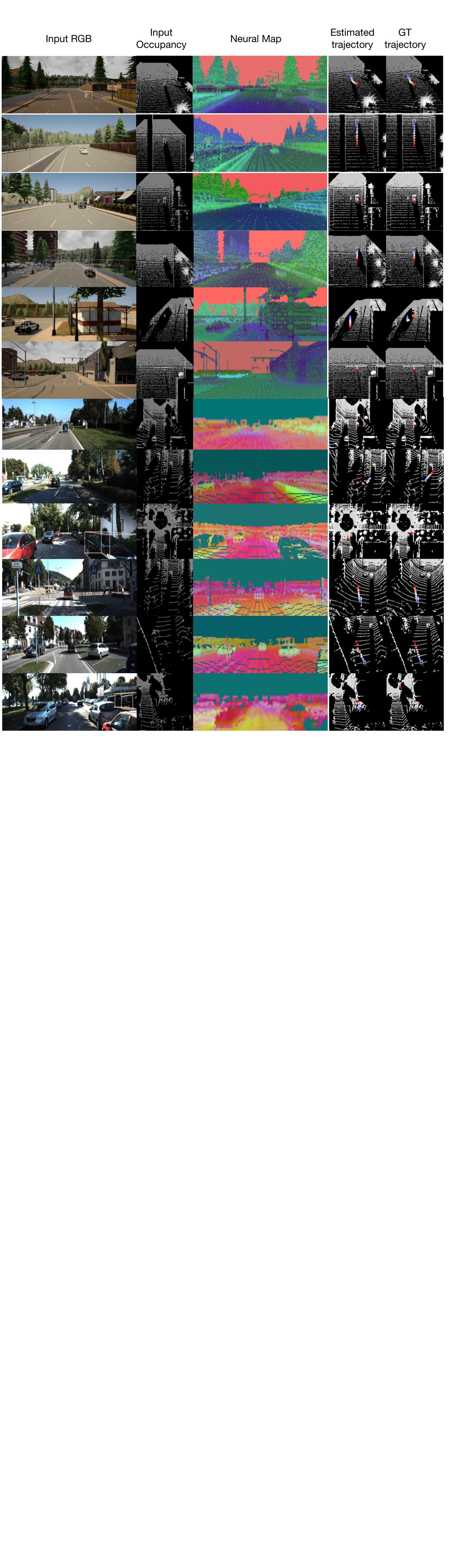}
  \caption{ \textbf{Self-supervised 3D object tracking}. In the 1st and 2nd column, we visualize the RGB, the object to track and depth from the first time frame, which is given as input to our model. In the 3rd column, we visualize our inferred point features by projecting them to the same RGB image and then doing PCA compression. In the last two columns, we show the estimated and ground truth trajectories. The top six rows show our results on CARLA; the
bottom six rows show our KITTI results.}
  \label{fig:tracking_quali_suppli}
\end{figure*}

    Figure \ref{fig:tracking_quali_suppli} shows more qualitative tracking results on CARLA and KITTI dataset. We have further attached videos of estimated and ground truth trajectories on our project page, and we urge the readers to have a look at those videos for a better understanding of the dataset complexity and our model's capabilities.  

\begin{figure*}[h!]
  \centering
  \includegraphics[width=0.72\textwidth]{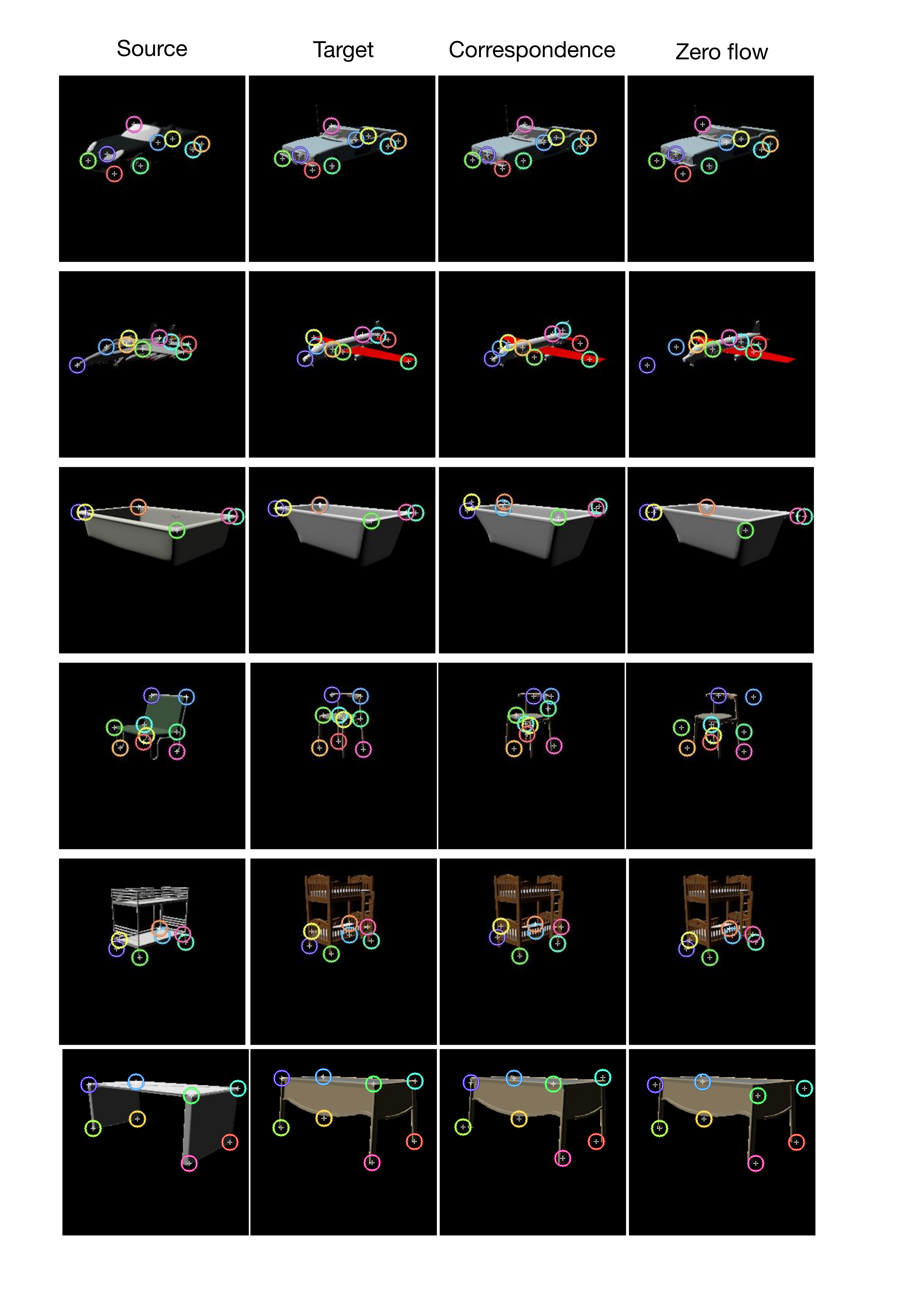}
  \caption{\textbf{Cross-object correspondence}. First column shows the Source entity. Second column shows the Target entity. The keypoints shown for these two columns are the ground truth keypoints. Third column shows the keypoints in Target entity inferred using correspondence from Source to Target. Last column shows the keypoint locations obtained assuming zero-flow correspondence from Source to Target, i.e. a point at location $(x, y, z)$ in Source gets mapped to the same location in Target.}
  \label{fig:correspred}
\end{figure*}

\subsection{Dense correspondence}
In this section, we evaluate our features on the task of establishing correspondence between two instances, $I_1$ (the Source entity) and $I_2$ (the Target entity), belonging to the same category. We use the keypoints provided by You \etal \cite{you2020keypointnet} on the ShapeNet dataset for qualitative evaluation. We assume that the input objects are already aligned. Given the dense pointcloud for both the objects, we featurize the points using the approach proposed in Section \ref{sec:model}. Then, for a point $(x, y, z)$ with feature $f_{x,y,z}$ in $I_1$, we find the corresponding point $(x', y', z')$ in $I_2$ whose feature $f'_{x',y',z'}$ has the highest cosine similarity with $f_{x,y,z}$. Since the instances are already aligned, we limit this search to a local neighborhood of radius 10cm. Figure~\ref{fig:correspred} shows the cross-object correspondence results achieved using the features learned by our model.

\subsection{Using only RGB for pose estimation}
\label{sec:poseestimation}
In this section, we show the comparison for cross-scene and cross-object 3D alignment, while operating our model on RGB versus RGB-D input. For this we follow the same experimental setup as of Section \ref{sec:corres_exp}. We refer our model operating on RGB as \model-RGB. We show the comparison with \model in Table \ref{tab:correspondence_rgb}.
\begin{table}[h!]
\centering
\begin{tabular}{ |c|c|c| } 
\hline
Method & cross-object & cross-view\\
\hline
\model{} & \textbf{0.18} & \textbf{0.58}\\
\model-RGB & 0.16 & 0.39\\ 
\hline
\end{tabular}
\caption{Cross-object and cross-view 3D alignment accuracies  in ShapeNet dataset (mean over 4 classes: Aeroplane, Mug, Car, Chair).}
\label{tab:correspondence_rgb}
\end{table}

\end{document}